%% file: main.tex
\documentclass[10pt,twocolumn,letterpaper]{article}

\usepackage{amsmath}
\usepackage{amssymb}   
\usepackage{pifont}    
\PassOptionsToPackage{hyphens}{url}

\usepackage[pagebackref,breaklinks,colorlinks,allcolors=myteal]{hyperref}

\usepackage[pagenumbers]{wacv} 

\usepackage[capitalize,noabbrev]{cleveref}
\crefname{section}{Sec.}{Secs.}
\Crefname{section}{Section}{Sections}
\Crefname{table}{Table}{Tables}
\crefname{table}{Tab.}{Tabs.}

\input{preamble}

\usepackage{graphicx}
\usepackage{amssymb}
\usepackage{booktabs}
\usepackage{bm}
\usepackage{tcolorbox}
\usepackage{caption}

\def\bmvaOneDot{.\,}

\def\eg{\emph{e.g}\bmvaOneDot}

\def\etal{\emph{et al}\bmvaOneDot}

\begin{document}

\title{Make me an Expert: Distilling from Generalist Black-Box Models into Specialized Models for Semantic Segmentation}
\author{Yasser Benigmim$^{1,2}$ \quad Subhankar Roy$^{3}$ \quad Khalid Oublal$^{1}$ \quad  Imad Eddine Marouf$^{1}$ \quad Slim Essid$^{4}$\thanks{This work was conducted while the author was at Télécom Paris.} \\ Vicky Kalogeiton$^{2}$ \quad Stéphane Lathuilière$^{5}$
    {}\\
    $^1$ LTCI, Télécom-Paris, Institut Polytechnique de Paris, France \\
    $^2$ LIX, Ecole Polytechnique, CNRS, Institut Polytechnique de Paris \\
    $^3$ University of Bergamo, Italy \quad
    $^4$ NVIDIA \quad
    $^5$ Inria at University Grenoble Alpes, LJK, France
}
\maketitle

\begin{abstract}
    \input{sec/0_abstract}
\end{abstract}
\input{sec/1_introduction}
\input{sec/2_related}

\input{sec/3_setting}
\input{sec/4_method}
\input{sec/5_experiments}
\input{sec/6_conclusion}

\bibliographystyle{ieee_fullname}
\bibliography{egbib}

\input{sec/x_supp}

\end{document}

%% file: preamble.tex
\PassOptionsToPackage{dvipsnames,table,xcdraw}{xcolor}
\usepackage{colortbl}
\usepackage{calc}
\usepackage{arydshln}
\usepackage{tikz}
\usepackage{algorithm}
\usepackage{algpseudocode}
\usepackage{amssymb}
\usepackage{booktabs}
\usepackage{makecell}
\usepackage{bbold}
\usepackage{multirow}
\usepackage{adjustbox}
\usepackage{enumitem}
\usepackage[normalem]{ulem}
\usepackage{pifont}
\usepackage{math}

\definecolor{OliveGreen}{RGB}{85,107,47}
\definecolor{Blue}{RGB}{0,0,255}
\definecolor{Red}{RGB}{255,0,0}
\definecolor{Orange}{RGB}{255,165,0}
\definecolor{lightblue}{RGB}{173,216,230}
\definecolor{lightyellow}{RGB}{255,255,224}
\definecolor{lightgreen}{RGB}{200,255,200}
\definecolor{lightorange}{RGB}{244,227,215}
\definecolor{darkgreen}{RGB}{0,100,0}
\definecolor{skyblue}{RGB}{135,206,235}

\newcommand{\miou}{mIoU\xspace}

\newcommand{\CLIPDINOISER}{CLIP-DINOiser\xspace}
\newcommand{\SAN}{SAN\xspace}

\definecolor{Blue}{HTML}{3A72B8}
\definecolor{Purple}{HTML}{9B70B6}
\definecolor{PurpleBlue}{HTML}{7264d8}
\definecolor{LightPurple}{HTML}{C5B6D8}
\definecolor{LightBlue}{HTML}{7EA6D6}
\definecolor{LightBlue}{HTML}{7EA6D6}
\definecolor{LightBluePlus}{HTML}{004a8f} 
\definecolor{GreenBlue}{HTML}{13B6AE}

\definecolor{Gray}{HTML}{A3A3A3}
\definecolor{lightgray}{HTML}{F0F0F0} 
\definecolor{LightGreen}{HTML}{d5f6f9} 

\definecolor{BlueTable}{HTML}{0087ff}
\definecolor{BlueTeal}{RGB}{0, 128, 128} 
\definecolor{MagentaBlack}{RGB}{139, 0, 139} 
\definecolor{myteal}{rgb}{0.0, 0.5, 0.5}

\definecolor{BlueTable}{HTML}{0087ff}
\definecolor{BlueTeal}{RGB}{0, 128, 128} 
\definecolor{MagentaBlack}{RGB}{139, 0, 139} 
\definecolor{myteal}{rgb}{0.0, 0.5, 0.5}
\definecolor{tablecolor}{RGB}{189, 252, 250}
\newcommand{\cmark}{\textcolor{blue}{\bm{$\checkmark$}}}           
\newcommand{\xmark}{\textcolor{magenta!60!black}{\bm{$\times$}}}              

\definecolor{myblue}{HTML}{27B0F5}

\newcommand*\circled[1]{%
  \tikz[baseline=(char.base)]{
    \node[fill=BlueTable!60!white, text=white, shape=circle, draw=white, inner sep=1.5pt] (char) {#1};
  }%
}

\usepackage{enumitem}
\usepackage{thmtools}
\usepackage{thm-restate}

\AtBeginDocument{\Crefname{assumption}{Assumption}{Assumptions}}

\def\1{\bm{1}}

\def\rs{{\textnormal{s}}}

\def\vf{{\bm{f}}}

\def\vv{{\bm{v}}}

\DeclareMathAlphabet{\mathsfit}{\encodingdefault}{\sfdefault}{m}{sl}
\SetMathAlphabet{\mathsfit}{bold}{\encodingdefault}{\sfdefault}{bx}{n}

\def\gE{{\mathcal{E}}}

\def\gT{{\mathcal{T}}}

\def\sN{{\mathbb{N}}}

\def\sR{{\mathbb{R}}}



%% file: sec/0_abstract.tex
The rise of Artificial Intelligence as a Service (AIaaS) democratizes access to pre-trained models via Application Programming Interfaces (APIs), but also raises a fundamental question: how can local models be effectively trained using black-box models that do not expose their weights, training data, or logits, a constraint in which current domain adaptation paradigms are impractical ? To address this challenge, we introduce the Black-Box Distillation (B$^2$D) setting, which enables local model adaptation under realistic constraints: (1) the API model is open-vocabulary and trained on large-scale general-purpose data, and (2) access is limited to one-hot predictions only. We identify that open-vocabulary models exhibit significant sensitivity to input resolution, with different object classes being segmented optimally at different scales, a limitation termed the ``curse of resolution''. Our method, ATtention-Guided sCaler (ATGC), addresses this challenge by leveraging DINOv2 attention maps to dynamically select optimal scales for black-box model inference. ATGC scores the attention maps with entropy to identify informative scales for pseudo-labelling, enabling effective distillation. Experiments demonstrate substantial improvements under black-box supervision across multiple datasets while requiring only one-hot API predictions. Our code is available at \url{https://github.com/yasserben/ATGC}.

%% file: sec/1_introduction.tex
\vspace{-0.3cm}
\section{Introduction}\label{sec:intro}
\begin{figure}
    \includegraphics[clip, trim=0.0cm 0.cm 17.6cm 0.0cm, width=0.49\textwidth]{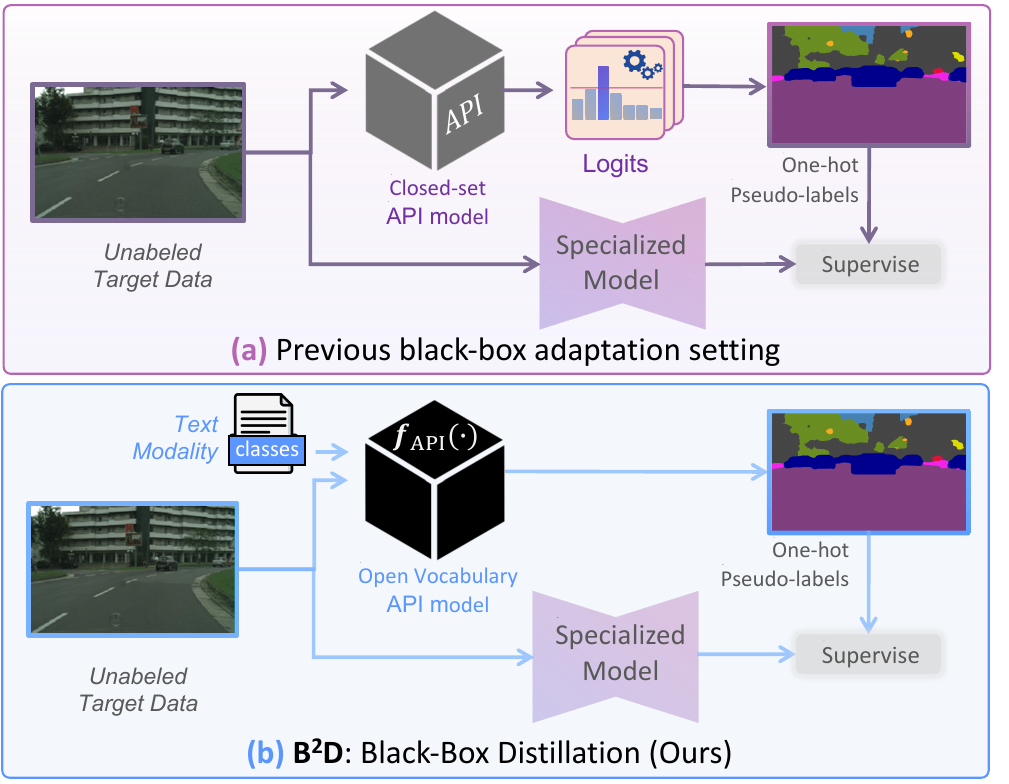}
    \centering
    \caption{\textbf{Comparison of black-box adaptation settings}. \textbf{(a)} Previous approaches assume access to API logits when leveraging pseudo-labels for student model training, which makes them ``gray-box''. \textbf{(b)} Our proposed \tasklongbb (\task) setting defines a more realistic ``black-box'' scenario, using open-vocabulary APIs without any access to logits.}
    \label{fig:teaser}
\end{figure}

The paradigm of pre-training neural networks on large datasets~\cite{tay-etal-2023-scaling, Cherti_2023_CVPR, bahri2024explaining} has produced powerful ``Foundation Models'' (FMs)~\cite{bommasani2021opportunities} with broad applicability across numerous domains~\cite{devlin2018bert,brown2020language,touvron2023llama, kirillov2023segment, liu2024visual, Yuan_2024_CVPR}. Many of these FMs are commercialized as Artificial Intelligence as a Service (AIaaS) and accessed through APIs, such as GPT-4~\cite{achiam2023gpt} and Gemini~\cite{comanici2025gemini}. While AIaaS simplifies infrastructure management, it presents users with significant challenges, including substantial costs at scale~\cite{chen2023frugalgpt}, network latency, and potential downtime.

\begin{figure*}[t]
    \includegraphics[clip, trim=0.0cm 0.cm 0.0cm 0.0cm, width=\textwidth]{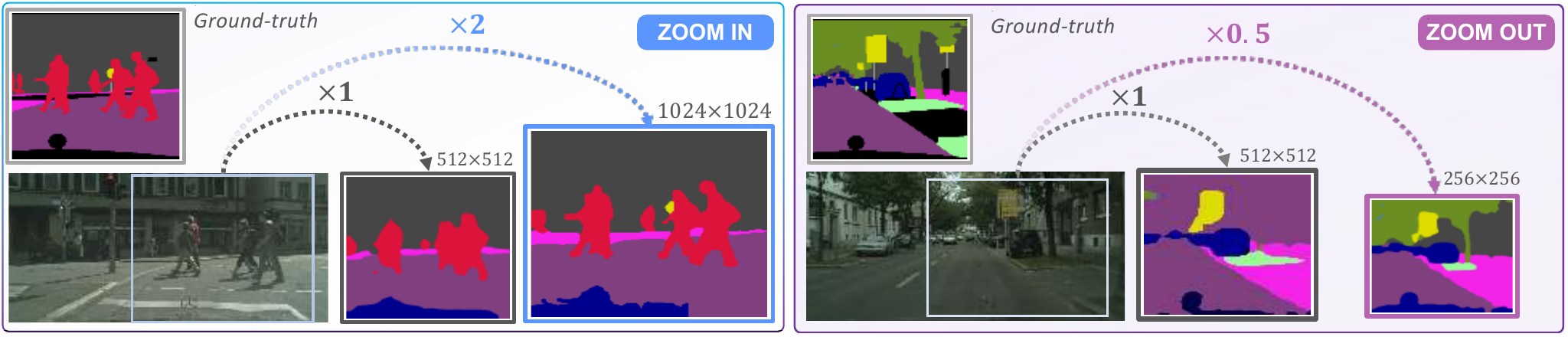}
    \centering
    \caption{\textbf{Scale-dependent segmentation quality}. We observe that segmentation performance varies with input resolution. Zooming in ($\times 2$, \textbf{left}) enhances segmentation of small, distant objects like pedestrians, while zooming out ($\times0.5$, \textbf{right}) improves large contextual elements by capturing broader spatial relationships. We refer to this issue as ``\textit{curse of resolution}'' where no single resolution optimally segments all object classes. This highlights the challenges inherent to distilling knowledge from open-vocabulary generalists models.}
    \label{fig:zoom}
    \vspace{-0.3cm}
\end{figure*}

A viable alternative is knowledge distillation (KD)~\cite{hinton2015distilling,sun2023dime}, where knowledge from a
generalist \textit{teacher} is distilled into a compact \textit{student} model. This creates a local ``expert''\footnote{Our usage of the term ``expert'' differs from the KD literature, where expert model refers to the teacher, but closer to mixture of experts~\cite{jacobs1991adaptive}.} model
that is cost-effective by eliminating calls to external APIs and can be finetuned for downstream tasks. Such models can be created through standard KD if the teacher is \textit{open-weight}~\cite{hoffman2016fcns}, or via black-box adaptation for \textit{closed-weight} teachers accessible only through an API~\cite{cuttano2023cross}. We focus on the latter as it is both more challenging since it excludes techniques like weight initialization or intermediate layer distillation~\cite{yang2022knowledge,jiao-etal-2020-tinybert}, and more realistic given the growing prevalence of commercial API-based foundation models.

Existing black-box adaptation methods often leverage pseudo-labels (PLs) from an API model while assuming access to its output logits~\cite{zhang2021unsupervised,cuttano2023cross}. While these approaches are termed ``black-box'', we believe that ``gray-box'' (see \hyperref[fig:teaser]
{\cref{fig:teaser}(a)}) would be a more appropriate description, given the access to logits. The ``gray-box'' assumption is a notable limitation, as commercial APIs increasingly withhold logits to protect intellectual property. This makes truly black-box approaches, which have seen recent interest for Large Language Models (LLMs)~\cite{ho2022large,liu2023tinygsm,mukherjee2023orca,sun2024bbox}, more relevant.

With a similar goal in mind, in this work, we formalize and study the problem of distilling a generalist black-box API-based foundation model into an expert model, but for the task of semantic segmentation. In this new setting, what we call \tasklongbb (\task), we assume that the API model is truly black-box and serves an open-vocabulary semantic segmentation model, which we can prompt using natural language~\cite{xu2023open,jose2025dinov2} to extract pixel-level semantic labels. As shown in \hyperref[fig:teaser]{\cref{fig:teaser}(b)}, we use these PLs from the teacher to train the student model. The newly introduced \task is more challenging because: (i) strict black-box assumption limits the scope of PL filtering, and (ii) the resolution of the image crop presented to the API governs the quality of PLs (see \cref{fig:zoom}). Since the pre-training data and their statistics are never disclosed for strictly-private API models, querying the API with optimal resolution to obtain high-quality PL is non-trivial.

To tackle \task, we propose a student-teacher self-training framework designed to obtain higher quality PLs from the API model. In the absence of confidence thresholding, we employ a simple strategy of \textit{prompt ensembling}, where we prompt the API using an averaged embedding derived from various synonyms of a noun (\eg, ``train'', ``tram'', ``locomotive"). To determine the optimal image crop resolution for the API, we introduce a novel module, \methodlongbb (\method), that dynamically identifies the most suitable crop resolution. Our approach leverages the common finding in semantic segmentation literature that different object classes are best segmented at different image resolutions~\cite{hoyer2022hrda,chen2017deeplab}. For instance, ``distant traffic lights'' are better segmented with \textit{zoomed-in} crops, while ``closer trucks'' benefit from the larger context provided by \textit{zoomed-out} crops. We objectively assess the \textit{goodness} of a crop resolution using the attention maps of the student model, which we score using Shannon entropy. A lower entropy value indicates a more focused map, suggesting that the resolution is well-suited to the objects within the crop. To ensure meaningful attention maps, we initialize the student's encoder with an open-weight model (\eg, DINOv2~\cite{oquab2023dinov2}) and keep it frozen during training. \method then selects the crop yielding the lowest entropy score and feeds it to the API model to obtain the PLs. Our contributions can be summarized as follows:
\begin{itemize}
    \item We formalize \tasklongbb (\task), a novel setting which assumes no access to API logits in contrast to previous ``gray-box'' approaches, better reflecting the reality of commercial AI services.
    \item We propose \methodlongbb (\method), a novel model which uses DINOv2 attention maps to dynamically select optimal image resolutions for API queries, yielding higher-quality pseudo-labels.
\end{itemize}
We benchmark our framework on Cityscapes~\cite{cordts2016cityscapes} and ACDC~\cite{sakaridis2021acdc}, demonstrating superior performance over state-of-the-art methods. Our findings highlight both the challenges and
potential of specializing models from truly black-box APIs.

%% file: sec/2_related.tex
\section{Related work}
\label{sec:related}

We review the literature related to \task, but limit our discussion to works that tackle the downstream task of semantic segmentation, which is the scope of this work.\par
\vspace{1.5mm}
\noindent \textbf{Black-box adaptation} refers to adapting from a pretrained model to a new domain or task without accessing or modifying its internal parameters, treating the model as a ``black-box''. In semantic segmentation, the precursors to black-box adaptation include unsupervised domain adaptation (UDA), where both the source dataset and source model are available during adaptation~\cite{hoffman2016fcns,hoffman2018cycada,hong2018conditional,chang2019all,pan2020unsupervised}, and source-free UDA (SFUDA)~\cite{liu2021source,fleuret2021uncertainty,kundu2021generalize}, which assumes access to only the source model, but not the source data. Both of these settings can be considered as ``white-box'', as they allow access to the weights of the source model. Although white-box adaptation offers maximum flexibility and control, it also introduces several challenges such as security risks~\cite{tramer2017ensemble,liang2022dine} or weights being unavailable due to commercialization~\cite{achiam2023gpt}.

In contrast, black-box UDA (B$^{2}$UDA) discards the assumption of access to the source pretrained model’s weights, and instead treats it as a black box, accessible only via an API. B$^{2}$UDA has mainly been studied in the context of image classification~\cite{zhang2021unsupervised,liang2022dine,zhang2023black}, and has only recently been investigated for semantic segmentation~\cite{cuttano2023cross}. In detail, Cuttano \etal~\cite{cuttano2023cross} proposed a mechanism to extract reliable PLs from the black-box model through confidence-based filtering. A downside of this approach is that it involves accessing the logits from the API, which may not always be guaranteed, and thus can more accurately be considered as a ``gray-box'' approach~\cite{sun2024bbox}. Differently, in our proposed \task setting, we adhere to the true definition of a black-box setting -- neither the source model's weights nor the output logits are available. The key differences among the various settings have been summarized in Tab.~\ref{tab:setting_summary}.\par
\vspace{2mm}
\noindent\textbf{Distilling from foundation models}. Foundation models (FMs) are large-scale general-purpose models~\cite{radford2021learning,kirillov2023segment,oquab2023dinov2,awais2025foundation} trained on massive, diverse datasets, designed to serve as a solid ``foundation'' or starting point for various downstream tasks~\cite{awais2025foundation}. While the FMs offer many opportunities~\cite{bommasani2021opportunities}, their large memory footprint hinder deployment on resource-constrained devices~\cite{raspberrypi2019modelb}. To balance performance and efficiency, Knowledge Distillation (KD)~\cite{hinton2015distillingknowledgeneuralnetwork} has been adopted as a go-to technique to train a smaller student model using a FM as a teacher. At its core, KD-based approaches assume that the teacher is open-weight (\ie, white-box), and thus distill knowledge either using the teacher's output logits~\cite{mirzadeh2020improved,beyer2022knowledge,oquab2023dinov2}, internal feature representations~\cite{yim2017gift,you2017learning,zagoruyko2016paying,wei2022contrastive} or their combination~\cite{liu2020adaptive,sun2023dime}. Very recently, this idea has been extended to distill knowledge from multiple FMs into a single student model~\cite{ranzinger2024radio,sariyildiz2024unic,heinrich2025radiov2}, and has demonstrated strong performance in multiple tasks, including semantic segmentation. Despite the progress enabled by KD, the white-box (or gray-box) assumption makes it inapplicable in truly restrictive APIs. The \task setting proposed in this work is challenging as it allows only a ``hard'' variant of KD, through the use of one-hot PLs.

\begin{table}[!t]
    \centering
    \small
    \resizebox{\columnwidth}{!}{
        \begin{tabular}{l|c|cc|c}
            \toprule
            Settings                             & \makecell{Source                                                                                                                                                             \\data} & \multicolumn{2}{c|}{\makecell{Pretrained\\model}} & \makecell{Deployment\\feasibility} \\
            \midrule
            \textbf{UDA}~\cite{hoffman2016fcns}  & \cmark           & \includegraphics[width=0.04\columnwidth]{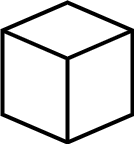} & \includegraphics[width=0.08\columnwidth]{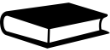} & Low      \\
            \textbf{SFUDA}~\cite{liu2021source}  & \xmark           & \includegraphics[width=0.04\columnwidth]{Figures/icons/white-box.png} & \includegraphics[width=0.08\columnwidth]{Figures/icons/closed-set.png} & Moderate \\
            \textbf{\bb}~\cite{cuttano2023cross} & \xmark           & \includegraphics[width=0.04\columnwidth]{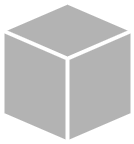}  & \includegraphics[width=0.08\columnwidth]{Figures/icons/closed-set.png} & Moderate \\
            \midrule
            \textbf{\task} (Ours)                & \xmark           & \includegraphics[width=0.04\columnwidth]{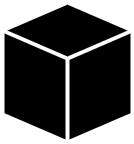} & \includegraphics[width=0.05\columnwidth]{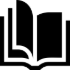}   & High     \\
            \bottomrule
        \end{tabular}
    }
    \caption{\textbf{Different model adaptation settings. }$\big($\includegraphics[width=0.04\columnwidth]{Figures/icons/white-box.png}, \includegraphics[width=0.04\columnwidth]{Figures/icons/gray-box.png}, \includegraphics[width=0.04\columnwidth]{Figures/icons/black-box.png}$\big)$ denote white-box, gray-box and black-box models, respectively. \bb methods assume access to logits, whereas our \task does not. $\big($\includegraphics[width=0.07\columnwidth]{Figures/icons/closed-set.png}, \includegraphics[width=0.04\columnwidth]{Figures/icons/open-set.png}$\big)$ denote closed-set and open-vocabulary pre-trained source model. Due to the use of open-vocabulary models, \task allows adaptation to any desired set of classes, making it highly flexible for deployment on downstream tasks. \vspace{-0.5cm}}
    \label{tab:setting_summary}
\end{table}

Unlike B$^{2}$UDA~\cite{cuttano2023cross}, which allows adaptation to a fixed set of categories (or vocabulary), distilling from open-vocabulary FMs is more appealing, as joint training with vision and language modalities allow for adaptation beyond a fixed vocabulary through flexible natural language prompts (\eg, \texttt{a photo of a [CLASS]}~\cite{radford2021learning,jose2025dinov2}), facilitating deployment of the student model to any domain/task of interest. However, this advantage may get eclipsed in certain scenarios when the target domain (\eg, medical images) exhibits a significant domain gap with respect to the FM pre-training dataset. This will lead to very noisy PLs~\cite{chattopadhyay2025zero}, especially since the pre-training dataset is often not disclosed by API providers. To balance flexibility and performance, we propose two strategies to extract more reliable PLs, while staying within the scope of the true black-box setting.

%% file: sec/3_setting.tex
\section{Problem formulation and preliminaries}
\label{sec:setting2}

The goal of \tasklong (\task) is to transfer knowledge from a \emph{generalist black-box} model, accessed via an API, to a \emph{local} model. Next, we formulate the problem and introduce some preliminaries.

\vspace{1mm}
\noindent\textbf{Problem setup.} We are interested in training a local model for the task of semantic segmentation. We assume that we have access to an API that serves an open-vocabulary segmentation model $\vf_\text{API}$, which can produce \textit{one-hot} semantic segmentation map when presented with an image $X \in \mathbb{R}^{3 \times h \times w}$. We denote the one-hot segmentation map as $\mathcal{M} = \{0,1\}^{\mid\mathcal{C}\mid \times h \times w}$, where $\mathcal{C} = \{c_1, c_2, \cdots, c_K\}$ is the set of $K$ class names (or vocabulary). Note, $\mathcal{C}$ is not a set of class indices, but strings (\eg, ``\emph{car}'', ``\emph{pedestrian}''), and is provided by the user.

Given a dataset of unlabelled images from the target domain, $\mathcal{D} = \{X_i\}^{n}_{i=1}$, our goal is to learn a local segmentation model $\mathcal{F}_\theta \colon X \mapsto [0,1]^{\mid\mathcal{C}\mid \times h \times w}$ that infers per-pixel class probabilities. The only supervision available for learning the parameters $\theta$ of $\mathcal{F}_\theta$, is the API model $\vf_\text{API}$.\par
\vspace{1.5mm}
\noindent\textbf{Challenges in \task.} While black-box adaptation is challenging in itself, \task is more challenging than \bb ~\cite{cuttano2023cross} in two key aspects: (\textbf{i}) \bb assumes that the API model was trained on a niche labelled source dataset (\eg, GTA~\cite{richter2016playing}) that has a relatively small domain gap with the target domain (\eg, Cityscapes). We argue that this strong assumption does not reflect real-world domain gaps, and as a result, the adaptation strategy may not generalize beyond academic benchmarks. Instead, in \task, we propose to leverage an open-vocabulary segmentation model as the API. This choice can be viewed as a ``double-edged sword'', since, unlike \bb, it offers the flexibility to distill information for any vocabulary, but at the same time may widen the domain gap when the target domain is very different from the (unknown) pre-training data distribution. (\textbf{ii}) Our setting operates under the stricter assumption that the API provides only one-hot segmentation maps ($\mathcal{M}$). This constraint is motivated by the practical limitations of many real-world APIs, which often return only final predictions. It was shown in~\cite{hinton2015distillingknowledgeneuralnetwork} that KD derives its strength from the use of ``soft targets'' alongside hard labels, preventing overfitting. Thus, the absence of privileged information in \task prevents the use of effective techniques, such as confidence-based pseudo-labelling~\cite{cuttano2023cross} or soft-dillation~\cite{sun2023dime}, to mitigate the impact of noisy PLs, making it a more challenging task.\par
\vspace{1.5mm}
\noindent\textbf{Preliminaries.} Knowledge distillation~\cite{hinton2015distillingknowledgeneuralnetwork} consists in training a smaller (student) model to mimic the behaviour of a larger (teacher) model. The idea is to compress the knowledge of the teacher into a lightweight, faster model, without losing much performance. This is achieved through a distillation loss $\mathcal{L}_\text{KD}$ that is a sum of KL-divergence loss, computed between the teacher's $p_t$, and the student's $p_s$ predicted probability distributions, and a cross-entropy loss between $p_s$ and the true hard label $Y$ of the sample:
\begin{equation}
    \label{eqn:kd}
    \mathcal{L}_\text{KD} = \alpha \text{KL}(p_t \Vert p_s) + (1 - \alpha) \mathcal{L}_\text{CE}(p_s, Y),
\end{equation}
where $\alpha$ is a hyperparameter. In \task, we do not have access to $p_t$, preventing us from directly using the formulation in Eq.~(\ref{eqn:kd}) to train $\mathcal{F}_\theta$.

%% file: sec/4_method.tex
\section{Methods}
\label{sec:method}

\begin{figure}[t]
    \centering
    \includegraphics[width=\columnwidth,keepaspectratio]{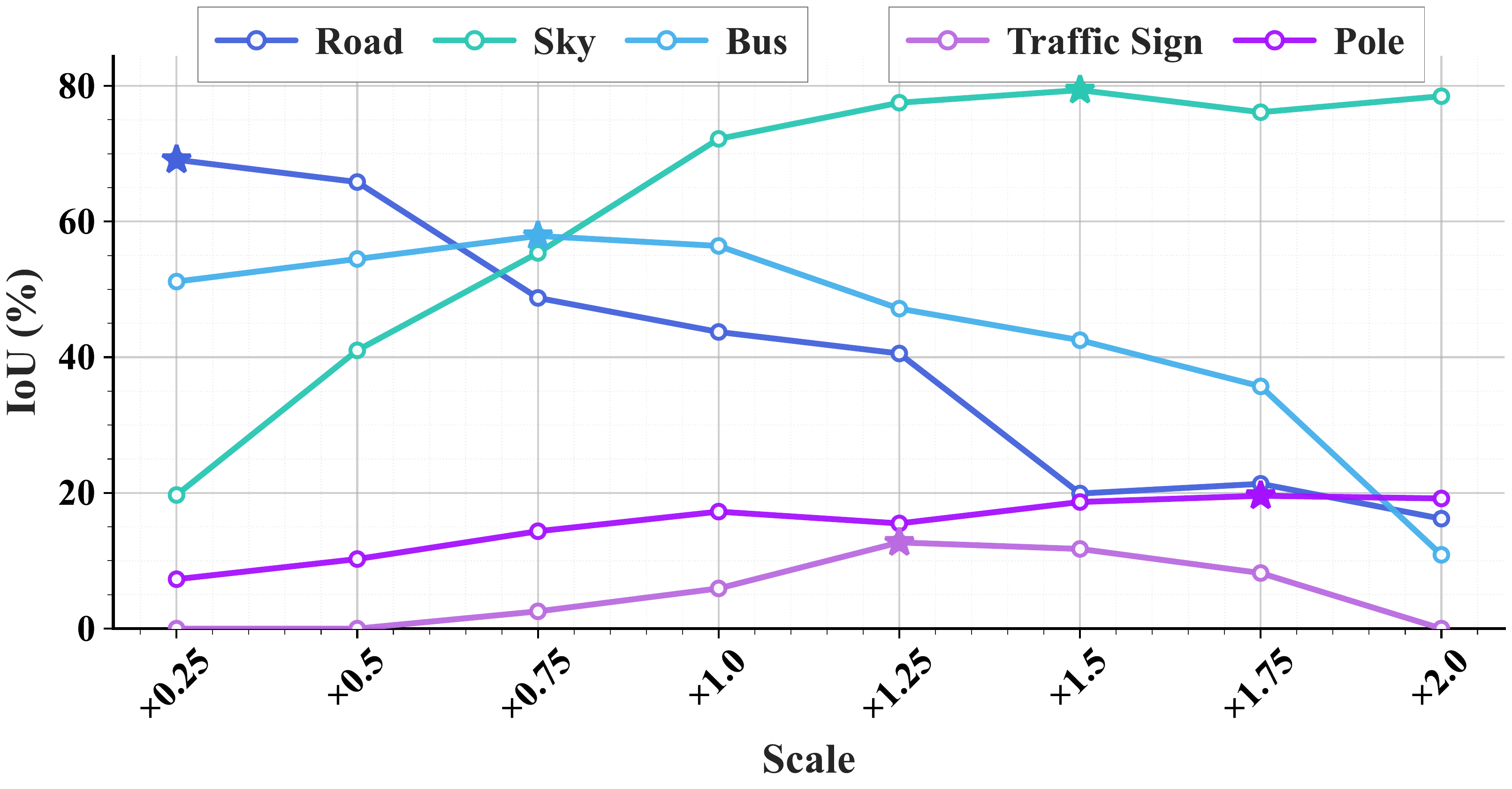}
    \caption{\textbf{Impact of scale on class-wise IoU performance. }The plot shows that performance varies across scales and across classes: larger-scale objects like ``road'' have peak performance at a lower resolution ($\times0.5$), while smaller-scale, distant objects like ``traffic sign'' is better segmented at higher resolutions ($\times1.75$).}
    \label{fig:effect_resolution}
\end{figure}
\noindent\textbf{Motivation.} In this work, we argue that the quality of the segmentation maps $\mathcal{M}$ derived from the API model $\vf_\text{API}$ is influenced by the resolution of the input image. To verify this, we conducted a preliminary study in which we varied the resolution (or scale\footnote{An image is scaled by an arbitrary factor, and crops of the size expected by the network are extracted using a sliding window. We use scale and resolution interchangeably.}) of Cityscapes images and fed them to an API model that serves an open-vocabulary segmentation model (SAN~\cite{xu2023side}). We prompted the API using the class names of Cityscapes and report in \cref{fig:effect_resolution} the class-wise intersection over union (IoU) performance for a selected few classes. From \cref{fig:effect_resolution} we observe that IoU for a class varies across scales, and there is no single scale that produces the best IoU across classes. For example, classes such as ``road'' and ``bus'' that cover a significant area of the scene require larger contextual information and are hence better segmented at zoomed-out resolution (or scale factors $< 1$). Conversely, some classes that are typically distant and cover a tiny area of the scene (\eg, ``traffic sign'' and ``pole'') show peak performance when zoomed in (or higher resolution, with scale factors $> 1$). This happens because small or distant objects do not require very long-range information to be well segmented. More qualitative and quantitative examples are provided in ~\cref{sec:varying-resolution,sec:effec-attention-maps}

\begin{figure*}[t]
    \includegraphics[clip, trim=0.0cm 0.cm 0.0cm 0.0cm, width=\textwidth]{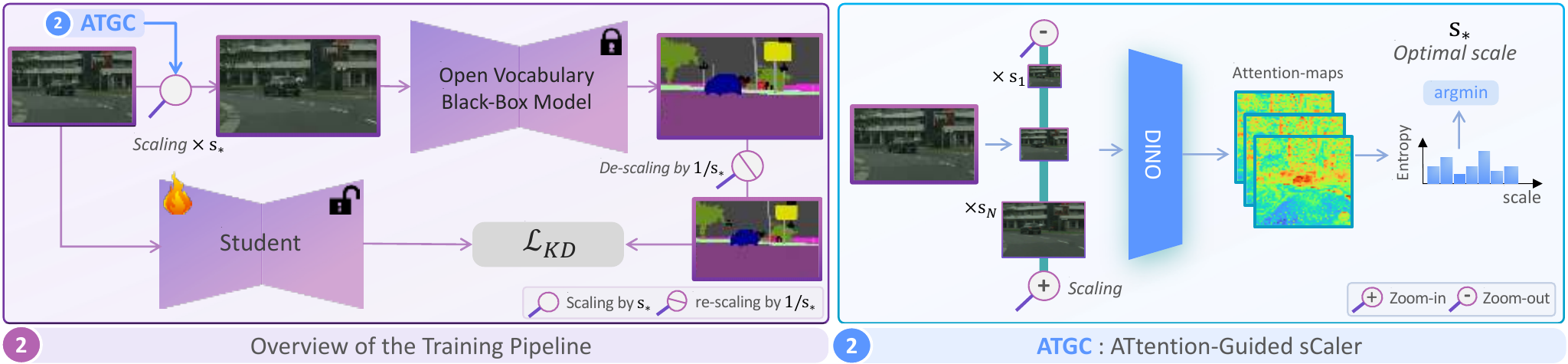}
    \centering
    \caption{
        \textbf{Overview of \method and our training pipeline.}
        (\textbf{Left}) \textcolor{gray}{\textbf{(1)}}~Training pipeline overview: Our approach operates under minimal access, relying solely on one-hot predictions $(0/1)$ from the open-vocabulary black-box model. The student network is trained using knowledge distillation loss $L_{KD}$ with pseudo-labels generated at optimal scales determined by our \method. (\textbf{Right}) \textcolor{gray}{\textbf{(2)}}~\method (\methodlongbb) detailed process: For each input image, multiple crops are generated and scaled to different resolutions. DINOv2 attention maps are computed for each scale, and entropy-based scoring identifies the optimal scale $S^*$ that yields the most informative features. The selected scale is used to query the black-box model, generating high-quality one-hot pseudo-labels that are re-scaled and used as supervision targets for student model training.
    }
    \vspace{-4mm}
    \label{fig:pipeline}
\end{figure*}

This phenomenon called \textit{the curse of resolution}, also noted in~\cite{hoyer2022hrda,zhu2024mrovseg}, is exacerbated when using an open-vocabulary segmentation model as the API, since the predominant resolution of its training images are likely to be different from the target dataset. The high variability in performance (\eg, IoU for the class ``road'' drops from $\sim70\%$ to $\sim20\%$ when scale is varied) clearly indicates that identifying an optimal resolution can significantly improve the quality of PLs, ultimately leading to a high-performing local segmentation model. This insight forms the foundation of our proposed method. Next, we give a brief overview of our method and then describe the novel component in detail.

\vspace{1mm}
\noindent\textbf{Method overview.} Our goal is to train a local segmentation model that becomes an ``expert'' at segmenting target images by using supervision from a ``generalist'' black-box API model, which in our case is an open-vocabulary segmentation model. We adopt a student-teacher framework~\cite{hoyer2022daformer,cuttano2023cross}, where the API model acts as a teacher and the local segmentation model acts as a student. The teacher supervises the student model by providing one-hot PLs to train its parameters. However, differently from~\cite{cuttano2023cross} we do not employ an exponential moving average (EMA) teacher derived from the student. Instead, we simply focus on extracting higher quality PLs to supervise the student.

\noindent As shown in \cref{fig:pipeline}, at the core of our student-teacher framework is a novel module, \method{} (\methodlongbb{}). We designed \method to identify (or \textit{mine}) the optimal scale that produces less noisy PLs than those obtained through naive approaches such as random scaling. \method exploits the attention maps of the student as a proxy to determine whether a given scale is suitable for segmenting the objects contained in a crop. We find that well-defined and crisp attention maps strongly correlate with superior dense prediction performance. Through a scoring mechanism, we pick the scale corresponding to the best attention map, and use that scale to preprocess the image crop and query the API to get PLs for supervision.

\subsection{\methodlongbb  (\method)}
\label{sec:cropper}

The proposed \method is a \textit{plug-and-play} module, applied on a given image crop before being fed to the API, whose goal is to choose the optimal scale. While a naive approach would be to perform random scaling (as in random resized crop data augmentation), we argue that it has two downsides: \textit{\textbf{(i)}} as discussed before, not all objects are well-segmented at every scale, and this will contribute to noisier PLs, which will be detrimental for the student when trained for a longer period~\cite{liu2020early}; and \textit{\textbf{(ii)}} it incurs unnecessary API calls, increasing cost due to inefficient queries. Therefore, \method facilitates learning in terms of both stable training and budget.

\noindent We design \method drawing inspirations from self-supervised learning methods~\cite{oquab2023dinov2,amir2022deep}, which have shown that Vision Transformers (ViTs) trained with self-supervision can yield strong localization properties. To exploit this property of ViTs we employ a pre-trained DINOv2~\cite{oquab2023dinov2} as the student encoder. In particular, we leverage the attention maps between the [\texttt{CLS}] token and the patch tokens from the final layer of DINOv2, and consider them a \textit{proxy indicator} of the suitability of a given image crop for pseudo-labelling. Next, we describe how we extract the attention maps, score them, and use the scores to find the optimal scale. We subsequently discuss why our design philosophy of utilizing attention maps works.\par
\vspace{1.5mm}
\noindent\textbf{Extracting Attention Maps.} Given an image crop $\Xmat \sim \mathcal{D}$, we scale it to $\sN$ different resolutions using a set of scale factors defining the support $\mathcal{S} = \{\rs_j \mid j \in \mathbb{N}, \rs_{min} \leq \rs_j \leq \rs_{max}\}$, where $\rs_{max}$, $\rs_{min} \in \sR^+$ are the maximum and minimum values.  We feed each scaled image $\Xmat_j = \gT_{\rs_j}(\Xmat)$ (where $\gT_{\rs_j} : \Xmat \in \sR^{3 \times h \times w} \mapsto \Xmat\in \sR^{3 \times h\cdot \rs_j \times w\cdot \rs_j } $ represents a transformation by scaling) to the DINOv2 feature extractor $\gE(\cdot)$ to get the attention maps from the final transformer block. Due to multi-headed attention, we get as many attention maps as the number of heads, which we average to get a single attention map per scale $s_j$ as $A_{j} \in \sR^{h_{j} \times w_{j}}$, where $h_{j}, w_{j}$ are the scaled spatial dimensions.

\vspace{1.5mm}
\noindent\textbf{Scoring Attention Maps.} To find the optimal scale $s^*$ for a given image crop $X$, we score the attention map $A_j$ with a metric $\mathbf{S}(\cdot)$ that can quantify the \textit{objectness} of the crop. To this end, we choose this metric to be Shannon entropy:

\begin{equation}
    \label{eqn:entropy}
    \mathbf{S}(A_j) = - \sum_{u,v} A_j(u,v) \log A_j(u,v),
\end{equation}
where the attention map $A_j$ is normalized to form a probability distribution over spatial locations $(u,v)$. We then select the scale $s_j$ that produces the lowest entropy attention map, as the optimal scale $s^*$:
\begin{equation}
    \label{eqn:optimal-scale}
    s^* = \arg\min_{s_j \in \mathcal{S}} \; \mathbf{S}(A_j)
\end{equation}

\begin{figure}
    \centering
    \includegraphics[clip, trim=0.0cm 0.cm 15.5cm 0.0cm, width=0.52\textwidth]{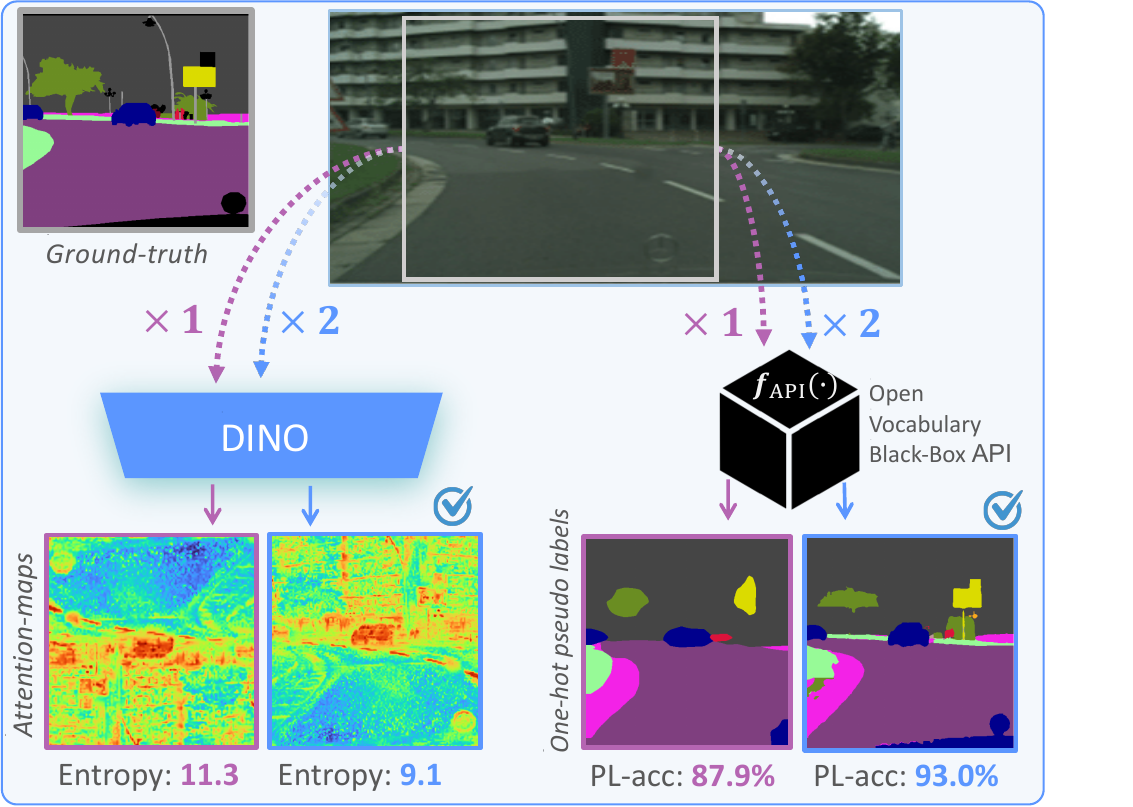}
    \caption{\textbf{DINOv2 and [\texttt{CLS}] Token for Optimal Resolution Selection.} DINOv2's [\texttt{CLS}] token attention maps are computed at multiple scales for each input image. The resolution with the highest spatially averaged attention score (\emph{e.g.}, $\times1$ vs. $\times2$) is selected to generate pseudo-labels from the API model.}
    \vspace{-4mm}
    \label{fig:attention-score-PL}
\end{figure}

The rationale is that an attention map with lower entropy exhibits peaked activations, signaling the presence of detected objects. In contrast, a higher entropy indicates a more diffuse attention map, suggesting that the model did not identify any particular semantic object. In \cref{fig:attention-score-PL} we visualize the attention-maps and the PLs and observe that a lower entropy attention map corresponds to better PL.\par
\vspace{1.5mm}
\noindent\textbf{Discussion.} A natural question arises: \textit{Why should the student encoder serve as a good proxy for a black-box API model?} At first glance, this may seem counterintuitive, given that the student encoder and the API model are two separate and independently trained entities. Viewing through the lens of the Platonic Representation Hypothesis~\cite{huh2024prh} -- which suggests that as models scale, they tend to converge toward learning the same underlying features -- we argue that both the API model and the student encoder (in our case, a pre-trained DINOv2), being trained on large-scale datasets, may develop internal representations that approximate the same statistical structures in representation space. Thus, both the models will react similarly to the same input, a property that we exploit in the absence of the access to the API's internal parameters. 

\subsection{Learning local segmentation model}
\label{sec:learn-student}
\noindent\textbf{Pseudo-labelling.} We use the optimal scale $s^*$ from Eq.~(\ref{eqn:optimal-scale}) to scale the input image crop as $X^{*} = \gT_{\rs^*}(X)$. We then feed $X^*$ to the API model to get the PL, denoted by $\hat{Y} = \vf_\text{API}\big(X^{*}, \mathcal{C}\big)$, where the vocabulary $\mathcal{C}$ are converted into standard text prompts~\cite{xu2023side,cho2024cat}, such as ``\texttt{a photo of a [CLASS]}''. More details are provided in \cref{sec:implem_details}

\noindent Although \method ensures better PL quality than using random scales, some pixels can still have erroneous PLs, causing the student to overfit on noisy PLs. To further improve the quality of supervision, we perform another round of PL-filtering at the pixel-level. In detail, we compute the agreement in class predictions (or PL accuracy) between the API and the student model. The PLs provided by the API for a given image are used to supervise the student only when the PL accuracy exceeds a predefined threshold $\tau$. We study the effect of varying $\tau$ in \cref{sec:supp_tau}.

\noindent\textbf{Student training.} We train the student model following Eq.~(\ref{eqn:kd}) by setting $\alpha=0$ and replacing $Y$ with $\hat{Y}$, which is equivalent to a cross-entropy loss with hard PLs. To preserve its rich internal representation, we train only the linear decoder and the last transformer block of the student model. Further implementation details are provided in \cref{sec:implem_details}.

%% file: sec/5_experiments.tex
\vspace{-4mm}
\section{Experiments}
\label{sec:experiments}
\subsection{Experimental setup}
\input{Tables/san_cityscapes}
\noindent \textbf{Implementation.} We conduct experiments on two unlabelled datasets following standard practices~\cite{wei2024stronger,hoyer2022hrda,hoyer2022daformer}. We use Cityscapes~\cite{cordts2016cityscapes} with $2975$ training and $500$ validation images at $2048\times1024$ resolution. We also use ACDC~\cite{sakaridis2021acdc} with $400$ training and $200$ validation images per weather condition (Night, Snow, Fog, Rain) at $1920\times1080$ resolution. We employ two open-vocabulary segmentation models as black-box models: \SAN~\cite{xu2023side} with a ViT-L/14 backbone trained on COCO-Stuff~\cite{caesar2018coco}, and \CLIPDINOISER, a training-free CLIP-based method with a ViT-B/16 backbone that has not been trained on segmentation data. For the local model, we use a DINOv2-S/14 encoder with registers~\cite{oquab2023dinov2} and a simple linear decoder.. We use thirteen different scale factors $\mathcal{S}$ = \{$0.25$, $0.28$, $0.34$, $0.38$, $0.44$, $0.47$,  $0.5$, $0.75$, $1.0$, $1.25$, $1.5$, $1.75$, $2.0$\} for resolution selection. Additional training details are provided in \cref{sec:implem_details}.

\vspace{1.5mm}
\noindent \textbf{Evaluation Protocol and Baselines. } We evaluate performance using mean Intersection over Union (\miou) across 19 classes for all datasets. Since this work introduces a new \task setting, no established methods exist for direct comparison. We therefore compare \method against CoRTE~\cite{cuttano2023cross}, the closest available approach. Since the official CoRTE code was not available, we reimplemented it for our experiments. We also include several baselines: (i) ``Naive Transfer" trains the student model using pseudo-labels from the API model at the default scale ($\rs_j\!=\!1$); (ii) ``Average" uses the average value of attention maps for scoring, with higher values preferred; (iii) ``Random" randomly selects a pseudo-label corresponding to one of the available scales for each image crop; (iv) ``Oracle" uses ground truth to evaluate all pseudo-labels at different scales and selects the one with highest pixel accuracy for training, providing an upper bound for our method; (v) ``Supervised" is trained with real ground truth labels. Note that ``Oracle" differs from ``Supervised" : Oracle uses the best API-generated pseudo-labels (selected using ground truth), while ``Supervised" uses actual ground truth labels for training.

\subsection{Main results}
\label{sec:main-results}
\input{Tables/clipdino_cityscapes}
\noindent \textbf{Specialization to Cityscapes. }\noindent We evaluate \method using two different open-vocabulary models as APIs: SAN~\cite{xu2023side} and CLIP-DINOiser~\cite{wysoczanska2024clip}. \cref{table:san_cityscapes,table:clipdino_cityscapes} show the results for both settings, respectively. For SAN as the API model, our method achieves 50.1 mIoU, outperforming all baselines including CoRTE$^{\dagger}$ (48.4), Naive Transfer (48.8), Random (48.6), and Average (44.6). With CLIP-DINOiser as the API, our method reaches 37.9 mIoU, again surpassing CoRTE$^{\dagger}$ (34.5), Naive Transfer (34.3), Average (34.1), and Random (35.5). Notably, our \method operates \textit{without} access to API logits, while CoRTE$^{\dagger}$ requires them, making our setting more realistic. Although our method falls short of Oracle performance (53.3 and 40.9 mIoU), it consistently outperforms existing approaches. The Oracle represents an upper bound achieved by always selecting optimal pseudo-labels, while the Supervised baseline (64.7 mIoU) shows potential performance when training with ground truth labels. These results validate our scale selection strategy's effectiveness for black-box model distillation.\par
\vspace{1mm}

\begin{figure}[ht]
    \centering
    \includegraphics[width=\columnwidth,keepaspectratio]{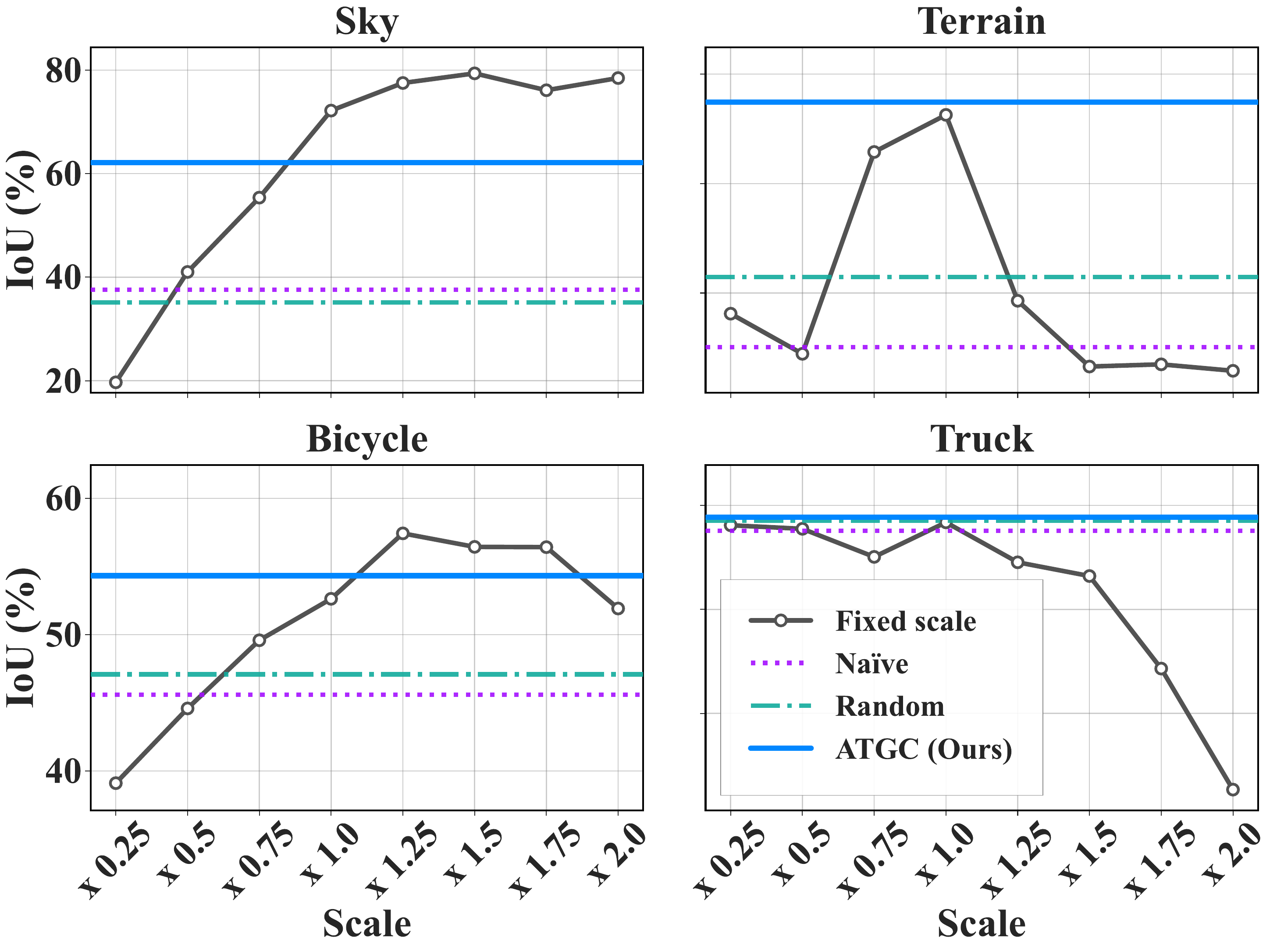}
    \caption{\textbf{Class-wise IoU performance. }The fixed scale approach ({\bf \color{gray} gray} line) shows performance when training at specific scales. At the same time, horizontal baselines represent  \textcolor{magenta!70!black}{\bf naive} (no scaling), \textcolor{GreenBlue!70!black}{\bf random} scale selection, and our \textcolor{BlueTable}{\bf ATGC} method. Results demonstrate that optimal scales vary significantly across classes, with ATGC consistently achieving competitive or superior performance compared to baseline approaches.}
    \vspace{-6mm}
    \label{fig:effect_scale}
\end{figure}

\input{Tables/acdc}

\noindent \textbf{Specialization to ACDC. } \cref{table:acdc} demonstrates \method's superior performance across both evaluation settings and black-box API models (SAN and CLIP-DINOiser). In the standard setting where models train on unlabelled ACDC images, \method achieves 41.0 mIoU with SAN and 25.0 mIoU with CLIP-DINOiser, consistently outperforming all baselines. Remarkably, in the domain generalization scenario, models trained on unlabelled Cityscapes and evaluated on ACDC achieve superior performance compared to direct ACDC training, a phenomenon we examine in \circled{3}.

\subsection{In-depth analysis}
\label{sec:depth}

\circled{1} \textbf{Does \method always find the ``optimal'' scales?.} To investigate this, we perform a grid search over several scales, where each scale is used to train a separate student model for the entire training duration. We then compare the performance of these fixed-scale models against \method. As reported in \cref{fig:effect_scale}, we observe that for the ``sky'' class, the performance of a student trained using PLs from a fixed scale of $\times1.5$ surpasses that of \method. Similarly, class ``bicycle'' benefits from scale $\times1.25$, a scale that \method fails to always find. In contrast, for certain classes such as ``terrain'' and ``truck'', \method achieves comparable or even superior results compared with the grid search scales. These results indicate that \method does not universally find the optimal scales for all the classes, which could be attributed to the presence of multiple competing classes within an image crop. However, \method is more compute-efficient and grid search is infeasible in practice due to the lack of any labelled data.\par
\vspace{2mm}
\noindent\circled{2} \textbf{How does the API model's training data affect performance?.} The quality of pseudo-labels, and thus student performance, is greatly dependent on the domain gap between the API model's training data and the target domain~\cite{roy2021curriculum}. This is evidenced by the performance gap between the original CoRTE results (55.5 mIoU on Cityscapes)~\cite{cuttano2023cross}, which used an API model trained on the closely-aligned GTA dataset, and our CoRTE$^{\dagger}$ baseline using SAN as an open-vocabulary API (48.4 in \cref{table:san_cityscapes}). This discrepancy arises because Cityscapes shares more similarities in scene layout and semantic classes with GTA than with COCO-Stuff, on which SAN was trained. The performance drop is even more pronounced with CLIP-DINOiser as the API, where CoRTE$^{\dagger}$'s performance decreases to 34.5 in \cref{table:clipdino_cityscapes}, as it is a training-free model that has not been exposed to any segmentation dataset. We argue that achieving high performance as in CoRTE's original setting can be illusory, as finding a target-aligned API model is often difficult in practice, particularly since the training data of commercial APIs is typically not disclosed. Therefore, assuming the API model is a generalist open-vocabulary model that covers a wide variety of target domains aligns better with real-world constraints.\par
\vspace{2mm}
\noindent\circled{3} \textbf{Does dataset size play a role?} We investigate why the DG results reported in \cref{table:acdc} consistently outperform direct training on ACDC, an observation that is counterintuitive in the DG literature~\cite{csurka2022semantic}. We believe that this difference is attributed to the smaller size of the ACDC dataset (2,975 in Cityscapes versus 1,600 in ACDC), as also shown in~\cite{lanzillotta2024testing}. Other factors could be at play, such as weather-corrupted images in ACDC, which are typically long-tailed in open-vocabulary pre-training datasets, resulting in noisier PLs. Thus, we conclude that it is important to have a sufficiently large dataset for effective \task, and caution must be exercised when working with target domains that are characterized by heavy corruptions or very long-tailed distributions.

%% file: Tables/san_cityscapes.tex
\begin{table*}[ht!]
\centering
\resizebox{\textwidth}{!}{%
\begin{tabular}{l|c|ccccccccccccccccccc|c}
\toprule
\textbf{Method} & \textbf{Logits} & \textbf{Road} & \textbf{SW} & \textbf{Buil.} & \textbf{Wall} & \textbf{Fence} & \textbf{Pole} & \textbf{TL} & \textbf{TS} & \textbf{Veg.} & \textbf{Ter.} & \textbf{Sky} & \textbf{PR} & \textbf{Rider} & \textbf{Car} & \textbf{Truck} & \textbf{Bus} & \textbf{Train} & \textbf{Mot.} & \textbf{Bike} & \textbf{mIoU} \\
\hline
\SAN~\cite{xu2023side}  & - &  88.7 &  50.1 &  82.0 &  29.2 &  33.4 &  3.1 &  22.5 &  29.8 &  81.3 &  20.9 &  86.1 &  53.5 &  0.7 &  73.1 &  23.9 &  60.0 &  50.3 &  46.8 &  53.0 &  46.8 \\
\hline
CoRTE$^{\dagger}$\cite{cuttano2023cross} & \checkmark & 88.6 & 49.9 & \bf 82.4 & \bf 29.9 & 28.9 & 0.0 & 29.4 & 32.2 & 82.6 & 26.3 & 81.1 & 55.3 & 0.0 & 76.0 & 26.4 & 67.8 & 56.0 & 47.2 & 58.4 & 48.4 \\
Naive Transfer & $\times$ & \bf 89.2 & \bf 50.4 & 82.3 & 28.3 & 31.9 & 0.9 & 32.1 & 33.1 & 82.5 & \bf 28.8 & 82.7 & 54.8 & 0.0 & 75.1 & 25.0 & 67.7 & \bf 56.4 & 48.0 & 57.8 & 48.8 \\
Average & $\times$ & 88.4 & 45.6 & 80.5 & 25.0 & 27.7 & 0.3 & 25.9 & 28.2 & 80.4 & 21.6 & 81.5 & 51.9 & 0.0 & 74.3 & 24.3 & 62.4 & 29.6 & 44.7 & 55.6 & 44.6 \\
Random & $\times$ & 88.2 & 48.2 & 82.3 & 27.9 & 31.9 & 5.5 & 31.2 & 37.0 & 82.9 & 26.0 & 83.1 & 55.8 & 0.0 & 75.8 & 24.6 & 67.3 & 48.2 & 47.9 & 59.4 & 48.6 \\
\midrule
\rowcolor{LightGreen}\textbf{\method (Ours)} & $\times$ & 86.6 & 46.3 & \bf 82.4 & 26.2 & \bf 35.3 & \bf 10.5 & \bf 34.6 & \bf 41.7 & \bf 83.8 & 28.6 & \bf 84.7 & \bf 58.0 & 0.0 & \bf 76.6 & \bf 23.8 & \bf 67.9 & 52.2 & \bf 50.8 & \bf 61.3 & \bf 50.1 \\
\midrule
\rowcolor{lightgray}Oracle & - & 91.8 & 56.8 & 84.4 & 40.7 & 41.7 & 9.4 & 35.6 & 41.3 & 84.8 & 35.0 & 83.7 & 57.6 & 0.0 & 79.2 & 28.4 & 70.0 & 60.8 & 50.8 & 61.1 & 53.3 \\
\rowcolor{lightgray}Supervised & - & 95.9 & 71.3 & 86.0 & 52.2 & 50.1 & 26.4 & 38.5 & 52.2 & 86.2 & 55.6 & 82.2 & 63.5 & 42.9 & 88.2 & 76.5 & 78.5 & 70.3 & 51.5 & 61.3 & 64.7 \\
\bottomrule
\end{tabular}
}
\caption{\textbf{Specialization to Cityscapes with SAN. }The best score for each column is highlighted in \textbf{bold}. Results are obtained using prompt engineering and averaged over 3 random seeds.$^{\dagger}$ Reimplemented by us as the original code was not available.}
\label{table:san_cityscapes}
\end{table*}
\setlength{\tabcolsep}{7pt}

%% file: Tables/clipdino_cityscapes.tex
\renewcommand{\arraystretch}{1.05}
\begin{table*}[ht]
\centering
\resizebox{\textwidth}{!}{%
\begin{tabular}{l|c|ccccccccccccccccccc|c}
\toprule
\textbf{Method} & \textbf{Logits} & \textbf{Road} & \textbf{SW} & \textbf{Buil.} & \textbf{Wall} & \textbf{Fence} & \textbf{Pole} & \textbf{TL} & \textbf{TS} & \textbf{Veg.} & \textbf{Ter.} & \textbf{Sky} & \textbf{PR} & \textbf{Rider} & \textbf{Car} & \textbf{Truck} & \textbf{Bus} & \textbf{Train} & \textbf{Mot.} & \textbf{Bike} & \textbf{mIoU} \\
\hline
\CLIPDINOISER~\cite{wysoczanska2024clip} & - &  68.5 &  17.0 &  73.7 &  23.4 &  24.0 &  10.1 &  2.2 &  3.5 &  78.8 &  10.2 &  43.3 &  45.2 &  9.2 &  69.3 &  26.1 &  52.9 &  26.8 &  27.4 &  42.3 & 34.4 \\
\hline
CoRTE$^{\dagger}$~\cite{cuttano2023cross} & \checkmark & \bf 64.9 & 18.5 & 75.8 & 23.1 & 18.5 & 10.0 & 0.0 & 0.0 & 80.5 & 4.5 & 36.9 & 52.0 & 3.0 & 72.9 & 30.2 & 57.1 & 24.6 & \bf 35.8 & 47.2 & 34.5 \\
Naive Transfer & $\times$ & 64.0 & 17.8 & 75.2 & 21.4 & 21.4 & 10.2 & 0.0 & 0.0 & 80.6 & 5.0 & 37.6 & 49.7 & \bf 4.0 & 71.3 & 27.6 & 56.0 & 28.5 & 35.0 & 45.6 & 34.3 \\
Average & $\times$ & 65.5 & 18.3 & 72.2 & \bf 24.3 & \bf 25.3 & 8.4 & 0.0 & 0.0 & 76.4 & 8.6 & 25.6 & 48.8 & 2.4 & 67.9 & 27.2 & 55.9 & 44.5 & 32.5 & 43.5 & 34.1 \\
Random & $\times$ & 64.8 & \bf 18.7 & 74.5 & 23.9 & 24.3 & 10.3 & 0.0 & 0.0 & 79.1 & 11.5 & 35.6 & 51.2 & 1.6 & 70.6 & 28.2 & \bf 58.2 & \bf 41.4 & 33.3 & 47.3 & 35.5 \\
\midrule
\rowcolor{LightGreen}\textbf{\method (Ours)} & $\times$ & 47.7 & 15.6 & \bf 78.0 & 22.5 & 24.2 & \bf 16.3 & \bf 0.2 & \bf 4.7 & \bf 83.1 & \bf 27.4 & \bf 62.1 & \bf 53.0 & 0.7 & \bf 73.1 & \bf 28.9 & 56.1 & 35.8 & 35.5 & \bf 54.3 & \textbf{37.9} \\
\midrule
\rowcolor{lightgray}Oracle & - & 76.7 & 24.4 & 78.8 & 30.5 & 31.3 & 14.2 & 0.0 & 2.5 & 81.5 & 25.4 & 49.3 & 52.8 & 6.5 & 74.1 & 33.4 & 62.3 & 48.6 & 34.1 & 49.9 & 40.9 \\
\rowcolor{lightgray}Supervised & - & 95.9 & 71.3 & 86.0 & 52.2 & 50.1 & 26.4 & 38.5 & 52.2 & 86.2 & 55.6 & 82.2 & 63.5 & 42.9 & 88.2 & 76.5 & 78.5 & 70.3 & 51.5 & 61.3 & 64.7 \\
\bottomrule
\end{tabular}
}
\caption{\textbf{Specialization to Cityscapes with CLIP-DINOiser. }The best score for each column is highlighted in \textbf{bold}. Results are obtained using prompt engineering and averaged over 3 random seeds.$^{\dagger}$ Reimplemented by us as the original code was not available.}
\label{table:clipdino_cityscapes}
\vspace{1mm}
\end{table*}

%% file: Tables/acdc.tex
\begin{table}[t]
\centering
\resizebox{\linewidth}{!}{
\begin{tabular}{llccccc}
\toprule
& \textbf{Method} &  \textbf{rain} & \textbf{fog}  & \textbf{night} & \textbf{snow} & \textbf{Average} \\
\midrule
\multirow{10}{*}{\rotatebox{90}{\textbf{SAN}}} 
& \SAN~\cite{xu2023side} &  45.5 &  45.9 &  31.8 &  48.0 &  42.8 \\
\cmidrule{2-7}
& CoRTE~\cite{cuttano2023cross} & 42.7 & \bf 45.7 & 31.9 & 42.0 & 40.6 \\
& Naive Transfer & 40.9 & 45.3 & 32.0 & 41.2 & 39.9   \\
& Random  & 41.1 & 45.2 & 32.5 & 41.0 & 40.0 \\
& \cellcolor{LightGreen}\textbf{\method} & \cellcolor{LightGreen} \bf 42.9 & \cellcolor{LightGreen} 45.1 & \cellcolor{LightGreen} \bf 33.6 & \cellcolor{LightGreen} \bf 42.3 & \cellcolor{LightGreen} \bf 41.0\\
\cmidrule{2-7}
& \cellcolor{lightgray}Oracle & \cellcolor{lightgray} 45.2 & \cellcolor{lightgray} 49.1 & \cellcolor{lightgray} 36.6 & \cellcolor{lightgray} 44.8 & \cellcolor{lightgray} 44.0 \\
& \multicolumn{6}{c}{\textit{Domain Generalization Evaluation}} \\
& Naive Transfer (DG) & 46.1 & 47.2 & 33.3 & 47.0 & 43.4 \\
& Random (DG)  & 45.9 & 48.0 & 33.3 & 47.8 & 43.8 \\
& \cellcolor{LightGreen}\textbf{\method (DG)} & \cellcolor{LightGreen} \bf 48.0 & \cellcolor{LightGreen} \bf 49.4 & \cellcolor{LightGreen} \bf 34.3 &  \cellcolor{LightGreen} \bf 48.5 & \cellcolor{LightGreen} \textbf{45.1} \\
\midrule
\midrule
\multirow{10}{*}{\rotatebox{90}{\textbf{CLIP-DINOiser}}} 
& \CLIPDINOISER~\cite{wysoczanska2024clip} &  32.0 &  31.6 &  13.7 &  30.4 &  26.9  \\
\cmidrule{2-7}
& CoRTE~\cite{cuttano2023cross} & 24.9 & 24.5 & 10.8 & 26.6  & 21.7  \\
& Naive Transfer & 25.1 & 21.6 & 11.1 & 22.4 & 20.1  \\
& Random  & 26.1 & 24.0 & 12.5 & 23.4 & 21.5 \\
& \cellcolor{LightGreen}\textbf{\method} & \cellcolor{LightGreen} \bf 27.7 & 
\cellcolor{LightGreen} \bf 27.9 & 
\cellcolor{LightGreen} \bf 17.2 & 
\cellcolor{LightGreen} \bf 27.0 & 
\cellcolor{LightGreen} \bf 25.0  \\
\cmidrule{2-7}
& \cellcolor{lightgray}Oracle & \cellcolor{lightgray} 34.6 & \cellcolor{lightgray} 38.5 & \cellcolor{lightgray} 23.1 & \cellcolor{lightgray} 34.7 & \cellcolor{lightgray} 32.7 \\
& \multicolumn{6}{c}{\textit{Domain Generalization Evaluation}} \\
& Naive Transfer (DG) & 29.4 & 31.4 & 18.2 & 31.7 & 27.7 \\    
& Random (DG)  & 32.0 & 32.6 & 19.7 & 32.9 & 29.3 \\
& \cellcolor{LightGreen}\textbf{\method (DG)} & \cellcolor{LightGreen} \bf 36.8 & \cellcolor{LightGreen} \bf 37.2 & \cellcolor{LightGreen} \bf 23.6 & \cellcolor{LightGreen} \bf 39.7 & \cellcolor{LightGreen} \textbf{34.3} \\
\bottomrule
\end{tabular}
}
\caption{\textbf{Specialization to ACDC.} Results of training \textbf{\method} on the unlabelled ACDC dataset compared to state-of-the-art specialization methods across varying weather conditions. The best score for each column is highlighted in \textbf{bold}.}
\label{table:acdc}
\vspace{1.5mm}
\end{table}

%% file: sec/6_conclusion.tex
\section{Conclusions and Limitations}
\label{sec:conclusion}
In this work, we introduced \tasklong in semantic segmentation (\task), a realistic adaptation paradigm that operates with only one-hot predictions from open-vocabulary API models, removing impractical assumptions of existing black-box approaches. We identified the ``curse of resolution'', whereby different object classes achieve optimal segmentation at different input scales, and proposed \method (\methodlongbb), which leverages DINOv2 attention maps and entropy scoring to dynamically select optimal scales for black-box inference, demonstrating effectiveness in generating high-quality pseudo-labels.

Although our experiments show promising results, several limitations remain. As discussed in \cref{sec:depth}, \method operates at the image level rather than the class level and uses static prompts that do not adapt to varying object scales. This can produce noisier pseudo-labels when multiple classes benefiting from different scales appear in the same crop. Future work could address this by introducing class-level scaling or dynamic prompts. Additionally, our current approach does not account for API call budgets; efficiency could be improved by focusing on the most informative regions of an image to accelerate convergence.
\section*{Acknowledgements}
This work has been supported by the French National Research Agency (ANR) with the ANR-20-CE23-0027 project and was granted access to the HPC resources of IDRIS under the allocations AD011013071 and AD011014921 made by GENCI. We also thank Nacereddine Laddaoui for his invaluable help with infrastructure.

%% file: sec/x_supp.tex
\clearpage
\appendix
\setcounter{page}{1}
\maketitlesupplementary

In this supplementary material, we provide comprehensive details about the experimental results from the main paper,  including additional qualitative results. The supplementary material is organized as follows: In Section~\ref{sec:pseudo}, we present the pseudo-code for Attention Maps construction. In Section~\ref{sec:implem_details}, we provide additional details about the training process of \method. Section~\ref{sec:quantitative} presents detailed experimental results and some failure case analysis. Finally, in Section~\ref{sec:future} we discuss the limitations and future directions in \tasklong (\task).

\section{Pseudo-Code}
\label{sec:pseudo}
We present the pseudo-code for \method in \cref{alg:amt} and \cref{alg:training}. \Cref{alg:amt} details the offline construction of Attention Maps for the entire dataset, while \cref{alg:training} describes the training procedure of \method that leverages these attention maps for resolution mining and pseudo-label generation. Note that in our implementation, we construct the attention maps offline, and do it only once for the entire dataset because it does not depend on the API model. Offline computation speeds up the training process, as multiple forward passes with varying crop resolutions are not needed to mine the optimal resolution.
\begin{algorithm}[ht!]
    \caption{\textbf{Attention Maps Construction}}
    \label{alg:amt}
    \begin{algorithmic}[1]
        \Require Dataset $\mathcal{D} = \{X_i\}_{i=1}^n$ of size $n$, Scale factors $\mathcal{S} = \{\rs_j \mid j \in \mathbb{N}, \rs_{min} \leq \rs_j \leq \rs_{max}\}$
        \Ensure Dataset of Attention Maps $\{A_{i,j}\}_{i=1,j}^{n,|\mathcal{S}|}$
        \For{each image $X_i$ in $\mathcal{D}$}
        \For{each scale factor $\rs_j$ in $\mathcal{S}$}
        \State $X_{i,j} \gets \gT_{\rs_j}(X_i)$ {\Comment{\textcolor{blue}{Scale image by factor $\rs_j$}}}
        \State $A_{i,j} \gets \gE(X_{i,j})$ {\Comment{\textcolor{blue}{Get DINOv2 attention map}}}
        \State $A_{i,j} \gets \gT_{1/\rs_j}(A_{i,j})$ {\Comment{\textcolor{blue}{Rescale to original dimensions}}}
        \EndFor
        \EndFor\\
        \Return Dataset of attention maps $\{A_{i,j}\}_{i=1,j}^{n,|\mathcal{S}|}$
    \end{algorithmic}
\end{algorithm}

\begin{algorithm}[ht]
    \caption{\textbf{Training Procedure with \method}}
    \label{alg:training}
    \begin{algorithmic}[1]
        \Require Dataset $\mathcal{D}$, Dataset of Attention Maps $\{A_{i,j}\}_{i=1,j}^{n,|\mathcal{S}|}$, Class names $\mathcal{C}$, API model $\vf_\text{API}$, Student network $\mathcal{F}_\theta$, Threshold $\tau$
        \For{each iteration $t$}
        \State Sample image $X$ from $\mathcal{D}$
        \State $X_c, A_c \gets \texttt{RandomCrop}(X, A)$ {\Comment{\textcolor{blue}{Get fixed-size crops}}}
        \State $s^* \gets \operatorname*{argmin}_{j} \mathbf{S}(A_{c,j})$ {\Comment{\textcolor{blue}{Find optimal scale with lowest entropy}}}
        \State $X_c^* \gets \gT_{s^*}(X_c)$ {\Comment{\textcolor{blue}{Scale image by scale factor $s^*$}}}
        \State $\hat{Y} \gets \vf_\text{API}(X_c^*, \mathcal{C})$ {\Comment{\textcolor{blue}{Get API pseudo-labels}}}
        \State $\hat{Y} \gets \gT_{1/s^*}(\hat{Y})$ {\Comment{\textcolor{blue}{Rescale pseudo-labels to original crop size}}}
        \State $\tilde{Y} \gets \mathcal{F}_\theta(X_c)$ {\Comment{\textcolor{blue}{Get student prediction}}}
        \State $\text{IoU} \gets \texttt{IntersectionOverUnion}(\hat{Y}, \tilde{Y})$ {\Comment{\textcolor{blue}{Compute consistency}}}
        \If{$\text{IoU} \geq \tau$}
        \State Update $\theta$ using Eq.~(\ref{eqn:kd}) with $\hat{Y}$ as supervision
        \EndIf
        \EndFor\\
        \Return Updated student model $\mathcal{F}_\theta$
    \end{algorithmic}
\end{algorithm}

\begin{figure*}[t]
    \centering
    \includegraphics[width=\linewidth]{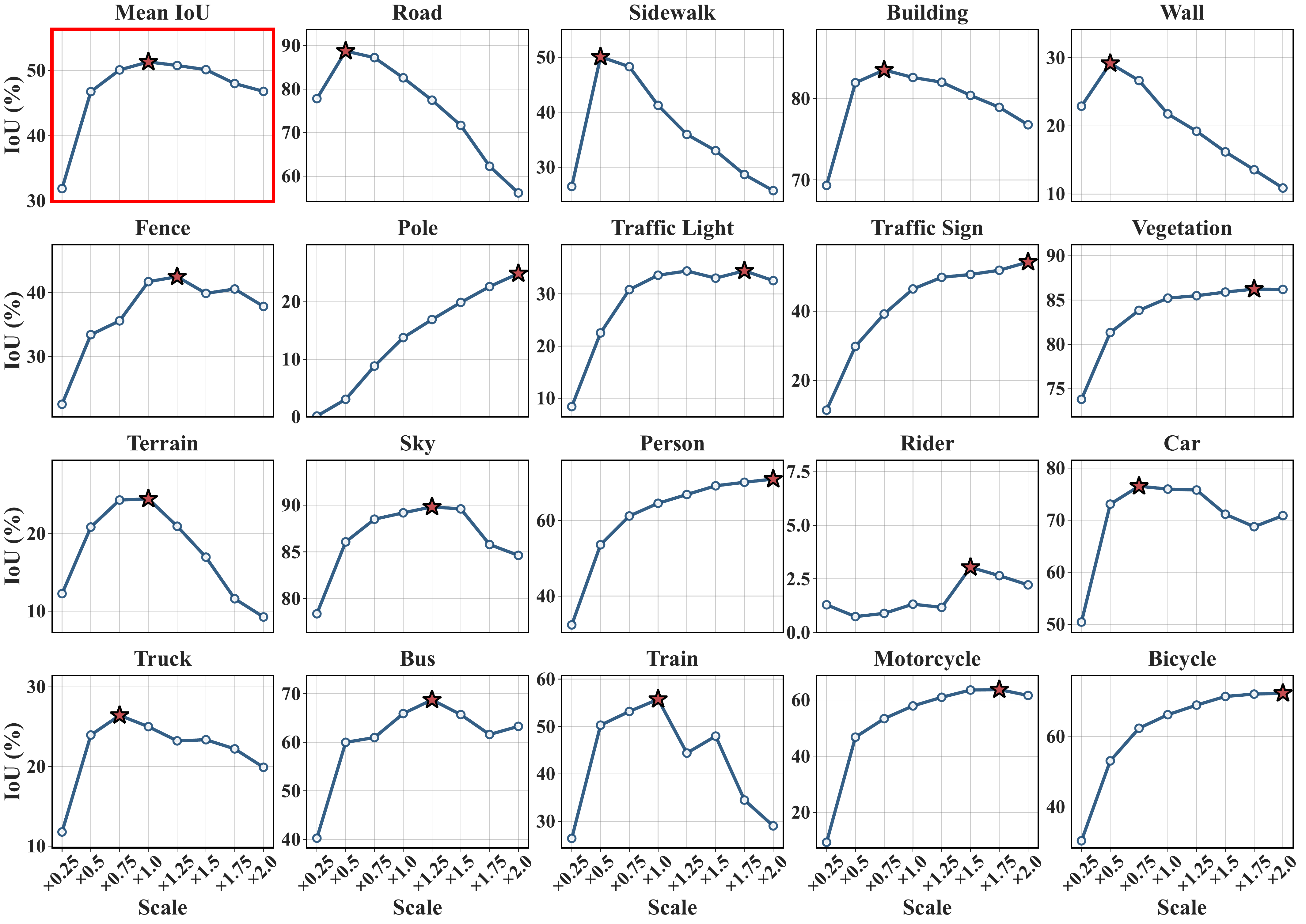}
    \caption{\textbf{Performance of SAN across all scaling factors.} The first subplot (top-left) shows the impact of resolution scaling on average mIoU, with subsequent subplots detailing individual class performances. Scaling factors ($\times0.25, \times1, ..., \times2$) represent resolution changes, yielding insights that highlight the optimal resolution for each class. \textcolor{Blue}{Markers} represent peak performance, highlighting the optimal resolution for each class. Small objects like ``pole'' and ``bicycle'' are better segmented in very high-resolution images which result in detailed crops given to the API model, while semantic classes such as ``road'' and ``wall'' are better segmented in low-resolution images yielding large context crops fed to the API model.}
    \label{fig:detailed_classes_perf_san}
\end{figure*}

\begin{figure*}[t]
    \centering
    \includegraphics[width=\linewidth]{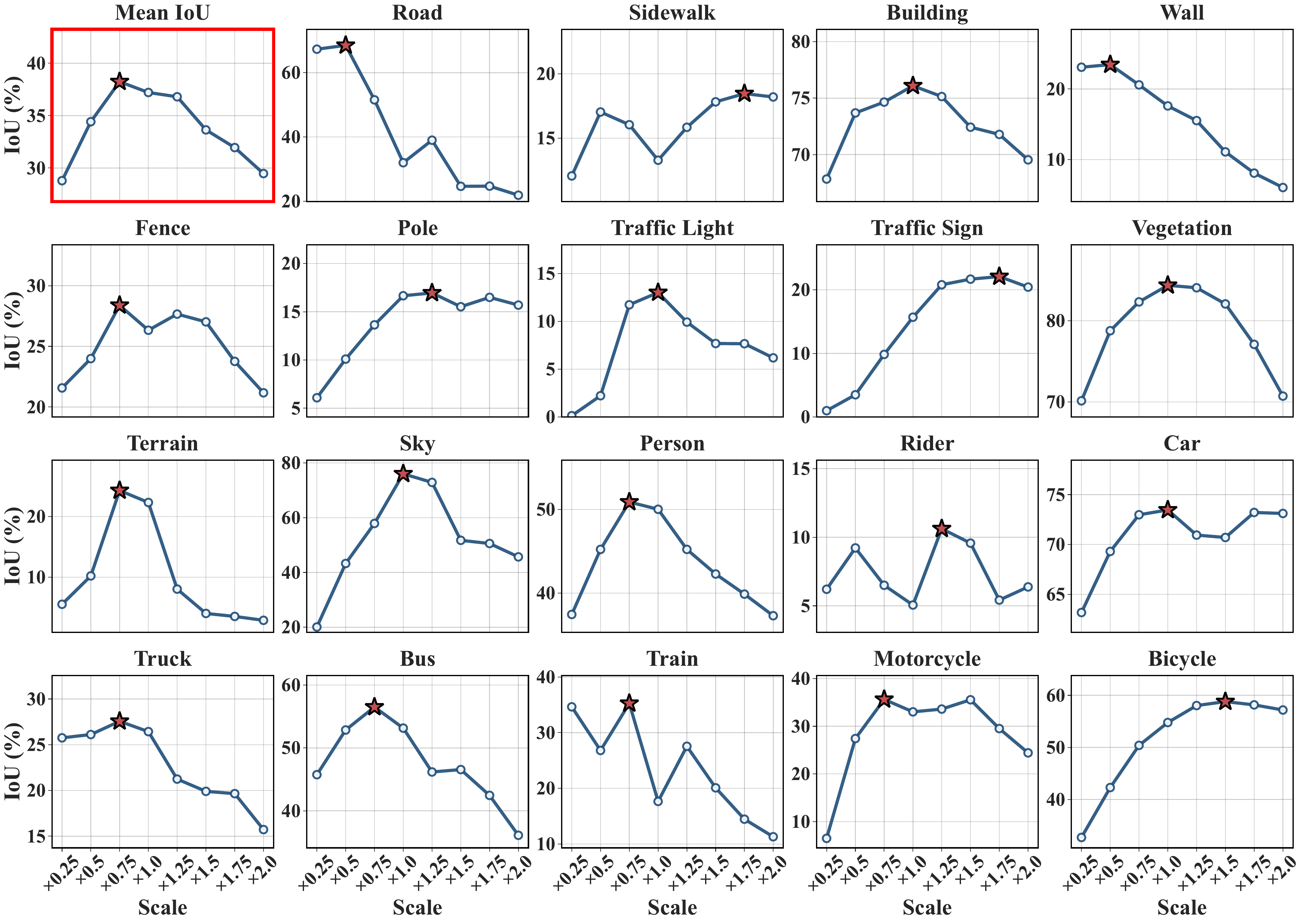}
    \caption{\textbf{Performance of CLIP-DINOiser across all scaling factors.} The first subplot (top-left) shows the impact of resolution scaling on average mIoU, with subsequent subplots detailing individual class performances. Scaling factors ($\times0.25, \times1, ..., \times2$) represent resolution changes, yielding insights that highlight the optimal resolution for each class. \textcolor{Blue}{Markers} represent peak performance, highlighting the optimal resolution for each class.}
    \label{fig:detailed_classes_perf_clipdino}
\end{figure*}

\begin{figure*}[t]
    \centering
    \includegraphics[width=\linewidth]{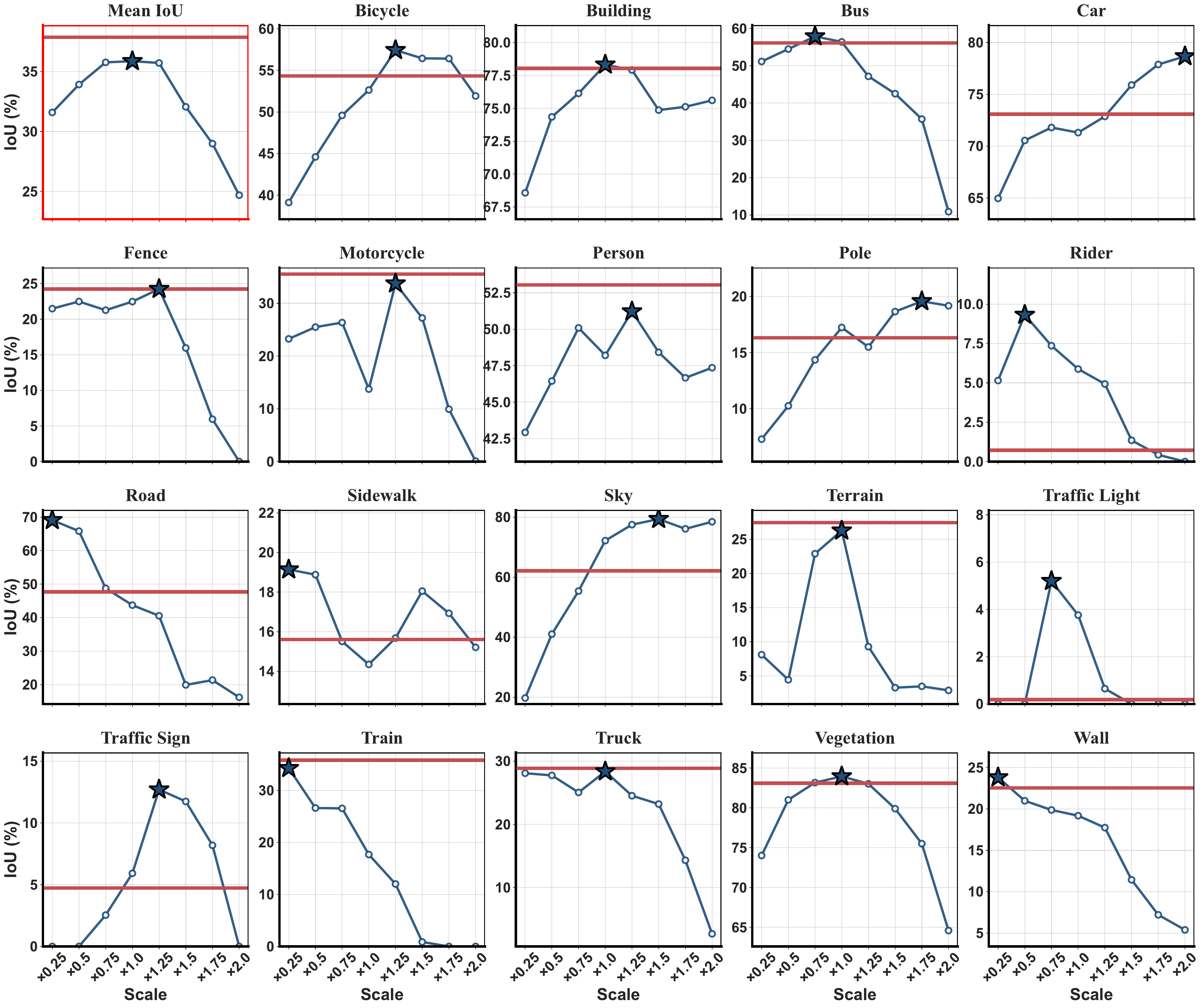}
    \caption{\textbf{Class-wise performance after training across all scaling factors.} The first subplot (top-left) shows the impact \textbf{training with a fixed resolution} on average mIoU, with subsequent subplots detailing individual class performances. The {\color{red}red} line represents our method {\color{red}\method}, and the {\color{blue} blue} curve shows results from training with fixed scaling factors. Each scaling factor ($\times0.25, \times1, ..., \times2$) indicates the scale applied to image crops during training. {\color{blue}Blue} markers highlight peak performance points, indicating the optimal scaling factor for each semantic class.}
    \label{fig:detailed_classes_ours}
\end{figure*}

\section{Implementation details}
\label{sec:implem_details}
We train our model with the AdamW optimizer~\cite{kingma2015adam} using a weight decay of 0.05 and a learning rate of $10^{-5}$ with polynomial decay. We set training iterations to 10K for Cityscapes and 6K for ACDC, with a batch size of 4.
\input{Tables/qualitatives_comparison}
Following standard evaluation protocols, the student model is evaluated on resized validation images to maintain consistent inference conditions across datasets, specifically resizing Cityscapes images from 2048×1024 to 1024×512, ACDC images from 1920×1080 to 960×540 for all weather conditions (Night, Snow, Fog, Rain). All reported performance metrics reflect the student model's capabilities after the full training schedule, ensuring fair comparisons with existing methods and maintaining computational efficiency.

\paragraph{Prompt engineering.} In this work, we have considered two open-vocabulary segmentation models --CLIP-DINOiser~\cite{wysoczanska2024clip} and SAN~\cite{xu2023side} -- as the API model. Following standard practice in open-vocabulary segmentation literature, we evaluate on 19 semantic classes from the Cityscapes dataset, using both primary class names and their synonyms to enhance model robustness (see class list below).
For prompting the open-vocabulary models, we used the commonly used 80 templates from CLIP for zero-shot image classification, commonly referred to as ImageNet templates ~\cite{radford2021learning}. We computed the average of the text embeddings from all templates to obtain the final text embedding for each class during inference.
\noindent\textbf{Classes:} [
    ``road, street, highway'',
    ``sidewalk, pavement, footpath'',
    ``building, structure, house'',
    ``wall, brick wall, stone wall'',
    ``fence, barrier, hedge'',
    ``pole, post, pillar'',
    ``traffic light, red light, green light'',
    ``traffic sign, stop sign, warning sign'',
    ``vegetation, plants, trees'',
    ``terrain, ground, grass'',
    ``sky, air, clouds'',
    ``person, pedestrian, people'',
    ``rider, biker, driver'',
    ``car, automobile, vehicle'',
    ``truck, pickup, van'',
    ``bus, shuttle, minibus'',
    ``train, tram, locomotive'',
    ``motorcycle, motorbike, scooter'',
    ``bicycle, bike, cycle'',
]

\section{Experimental results}
\label{sec:quantitative}
\subsection{Pseudo-label filtering.}
\label{sec:supp_tau}
\begin{figure}[ht!]
    \centering
    \includegraphics[width=\linewidth]{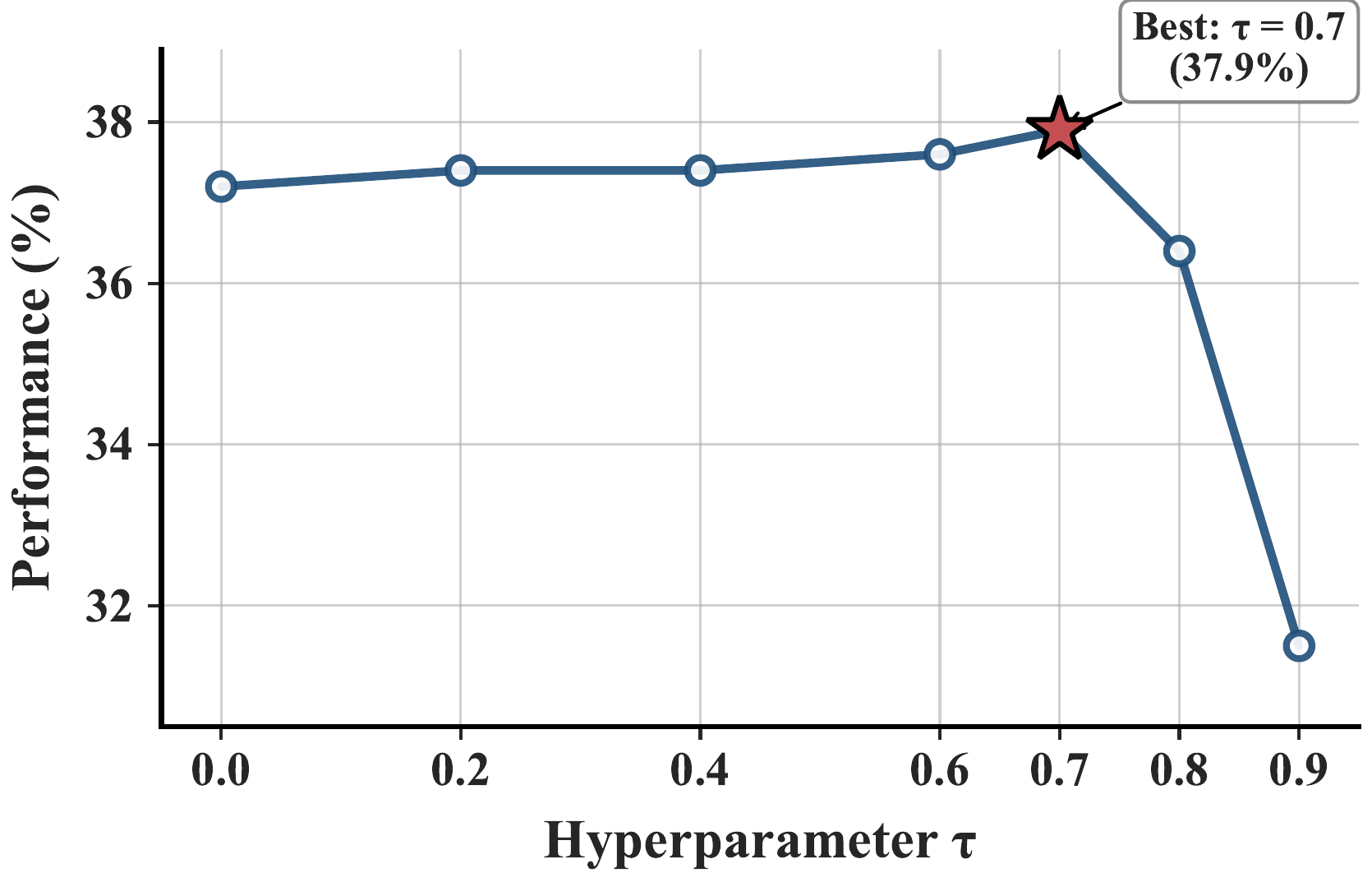}
    \caption{\textbf{Effect of confidence threshold $\tau$ on pseudo-label filtering performance. }}
    \label{fig:ablation_tau}
\end{figure}

We ablate the pseudo-label filtering threshold $\tau \in [0.0, 0.9]$ to study its impact on performance when using \CLIPDINOISER as the API model. As shown in \cref{fig:ablation_tau}, \method achieves optimal performance at $\tau = 0.7$ with 37.9\% mIoU.
Notably, \method demonstrates robust performance even when pseudo-label filtering is completely deactivated ($\tau\!=\!0.0$), achieving 37.2\% mIoU with only a modest 0.7\% decrease compared to the optimal threshold. This indicates that \method works effectively regardless of whether PL filtering is activated or not, showcasing the inherent quality of our scale-optimized pseudo-labels. Performance degrades more significantly when $\tau$ is too restrictive ($\tau\!=\!0.9$: 31.5\%), as overly strict filtering removes valuable training signals, reducing effective pseudo-label coverage. The optimal threshold $\tau\!=\!0.7$ achieves the right balance between pseudo-label quality and quantity. For consistency, we report all main results using $\tau\!=\!0.7$ throughout our experiments.

\subsection{Effect of varying resolutions}
\label{sec:varying-resolution}
\Cref{fig:detailed_classes_perf_san,fig:detailed_classes_perf_clipdino} show the performance of SAN and CLIP-DINOiser, respectively, when used as API models, across various scaling factors applied to the initial image resolution ($1024 \times 2048$) of the Cityscapes~\cite{cordts2016cityscapes} validation set. The subplots illustrate the effect of image resolution on segmentation accuracy for each semantic class and the average mIoU. Each subplot highlights scaling factors ranging from $0.25$ to $2.0$, with blue star indicating the optimal resolution for each class.

The first subplot shows that the average mIoU across all classes peaks at the scaling factor of $1.0$, reflecting the best overall performance. However, peak performance varies across classes at different scale factors. For example, classes such as ``person" and ``motorcycle" perform optimally at scaling factors of $2.0$ and $1.5$, respectively. In contrast, ``sidewalk" and ``truck" are better segmented at scaling factors of $0.5$ and $0.75$, respectively.

In general, the results emphasize that resolution scaling plays a critical role in achieving optimal performance for each class. Optimal resolution choices vary across different object types, with some benefiting from detailed crops taken from high resolution base images and others from large context crops taken from low resolution base images. This suggests that a tailored resolution strategy can improve segmentation outcomes using the API model, highlighting the importance of careful resolution scaling rather than adopting a one-size-fits-all approach.

\subsection{Qualitative comparison of methods}
In \cref{fig:qualitative_comparison}, we present the qualitative comparison of different methods. We have compared our proposed \method with a \bb approach CoRTe~\cite{cuttano2023cross}, and the Naive Transfer baseline that directly distills from the API model without considering varying resolutions. From the figure we observe that for larger object classes, such as ``car'' and ``bus'', and stuff classes, such as ``road'' and ``vegetation'' the segmentation quality is reasonably fair among all the baselines and our \method. However, when it comes to smaller objects, such as ``poles'' (highlighted with a white box in all the segmentation maps), the segmentation quality of our \method is much better. Furthermore, in row 3 of \cref{fig:qualitative_comparison} we observe in the highlighted region of the segmentation map that CoRTe is worse at segmenting ``sidewalk'' that is farther in the scene, compared to the nearby ``sidewalk''. Contrarily, both the Naive Transfer and \method is better at segmenting farther objects, with \method doing well on the farthest ``pole'' in the image. This observation underscores the need of obtaining pseudo-labels at the optimal crop resolution to reasonably segment objects that occupy a very small portion of the image.

\input{Tables/resolution_supp}

\subsection{Effectiveness of Attention Maps}
\label{sec:effec-attention-maps}
Building upon the motivational experiment in Sec. 4 and Fig. 4 of the main paper, which demonstrated a positive correlation between attention map quality (computed between DINO [\texttt{CLS}] and patch tokens) and segmentation performance, we present additional qualitative validation in \cref{fig:affinity-segmaps}.

These examples confirm that the relationship between attention map entropy and pseudo-label (PL) quality is consistent across different scenarios. Specifically, attention maps with lower entropy display concentrated, peaked activations that correspond to better object localization and, consequently, higher-quality pseudo-labels. Conversely, higher entropy values indicate spatially diffuse attention patterns that correlate with poor-quality pseudo-labels. Our experimental analysis validates that Shannon entropy of attention maps serves as an effective and reliable proxy for identifying the optimal scale for pseudo-label generation, which forms the core mechanism our method uses to distill knowledge from the API model.

\input{Tables/failure_supp}

\subsection{Failure cases}
In \cref{fig:failure_qualitative_comparison}, we report some failure modes of our \method, where we show that optimal crop resolution may not always result in the best pseudo-label predictions from the API model. In detail, for a given random crop, we visualize the prediction of the API model at the original crop resolution ($\xvect_c$) and the one with the optimal scale resolution ($\xvect_c^*$) as determined by the \method module. We observe from the figure that the quality of segmentation maps predicted by the API at the optimal scale is inferior when compared with the one obtained with the original crop resolution. Specifically, from \cref{fig:failure_qualitative_comparison}(c) and (d) we observe that the optimal crop resolution introduces strange artifacts in the segmentation maps, \eg, sharp discontinuities. Similar kind of observation holds for the other figures, where incorrect classes are predicted within other objects (\eg, \cref{fig:failure_qualitative_comparison}(a), where a ``truck'' pixels encroaches onto a ``car'' class pixels). This phenomenon occurs when extreme \textit{zoom-in} operations (with high scale factors) cause attention maps to be computed over homogeneous image regions, leading to highly concentrated activations for a single class and consequently producing misleadingly low entropy values.

Despite these pathological predictions, we employ a PL-filtering technique (as discussed in Sec. 4 of the main paper) that uses the consistency between the API predictions and the student model predictions to avoid training on erroneous PLs. Since the student model processes the image crop at original resolution, it is less likely to make the same errors as the API model that always generates PLs at the optimal scale resolution (see Fig. 3 of the main paper). We compute the pixel-level agreement between the API pseudo-labels and the student predictions using pixel accuracy, and only use pseudo-labels for supervision when this agreement exceeds a predefined threshold $\tau$. This filtering mechanism ensures that pathological API predictions are discarded and do not negatively impact the student network training.

\section{Limitations and future works}
\label{sec:future}
In this work, we introduced the task of \task, which distills knowledge from a generalist black-box model into a local segmentation model by mining optimal scales for querying the API. This approach represents a significant improvement over using default or random crop resolutions for pseudo-label generation. However, our current framework operates under the assumption that optimal scale selection is the primary determinant of pseudo-label quality. While our experiments validate that optimal scaling plays a crucial role in determining PL quality, several other factors may influence performance that remain unexplored, such as adaptive prompt engineering tailored to specific crop resolutions. Additionally, our framework lacks a systematic mechanism for selecting the most informative samples or image regions for API queries. This uniform treatment of all image regions is inefficient, particularly given the inherent class imbalance in driving datasets where ``stuff'' classes (road, sky, vegetation) occupy significantly larger pixel areas than ``things'' classes (vehicles, pedestrians, traffic signs).

As future work, we will explore other crucial factors that can further improve the quality of PLs and select most informative regions, thereby reducing the number of calls to the API, making black-box distillation from API calls more economically and computationally viable. Furthermore, we will also explore making API calls for pseudo-supervision while ensuring the privacy of sensitive local client data.

%% file: Tables/qualitatives_comparison.tex
\begin{table*}[t]
    \centering
    \begin{tabular}{ccccc}
       \begin{tikzpicture}
            \node[anchor=south west,inner sep=0] (image) at (0,0) {\includegraphics[width=0.18\textwidth]{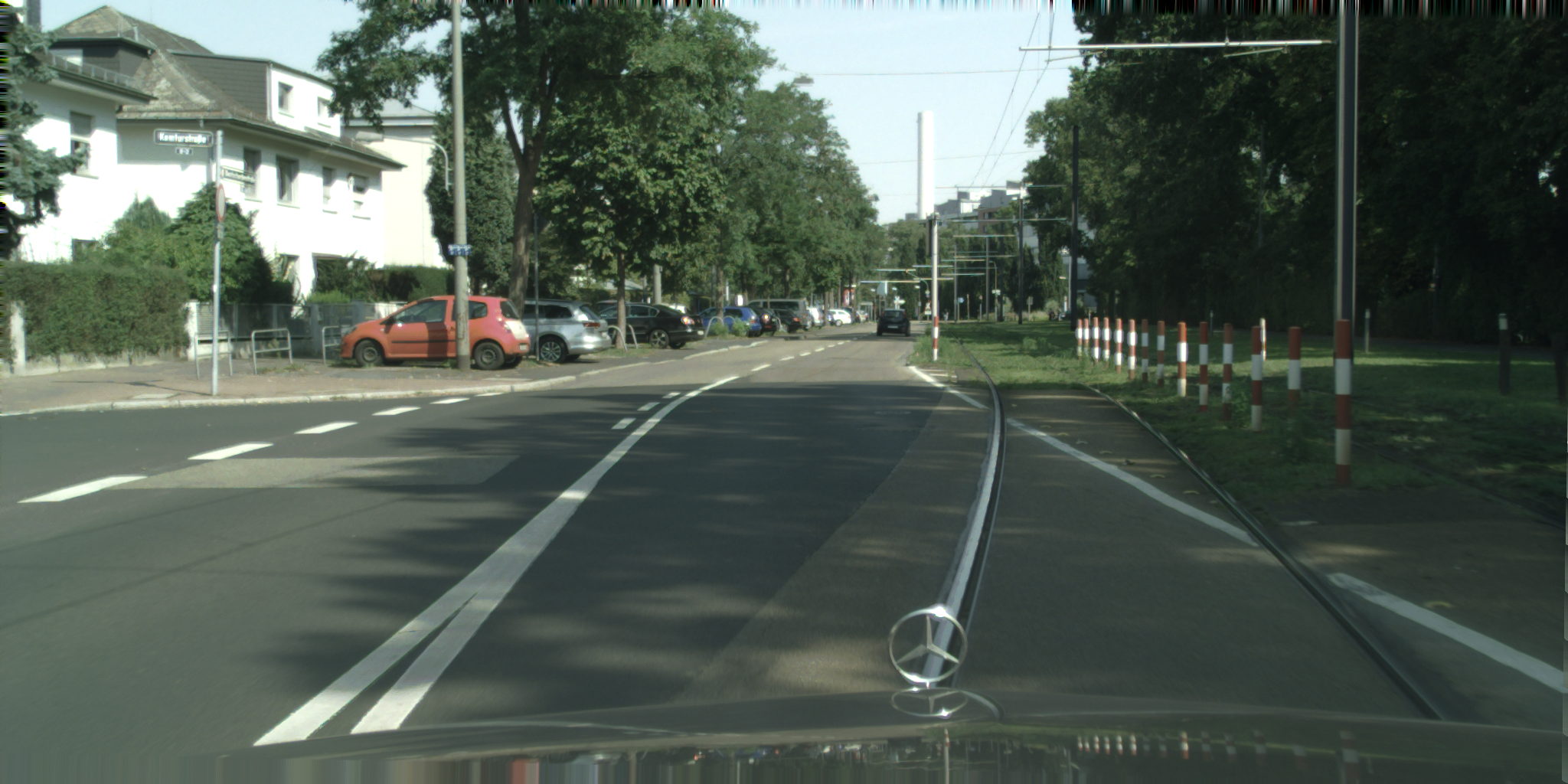}};
        \end{tikzpicture}  
        &
        \begin{tikzpicture}
            \node[anchor=south west,inner sep=0] (image) at (0,0) {\includegraphics[width=0.18\textwidth]{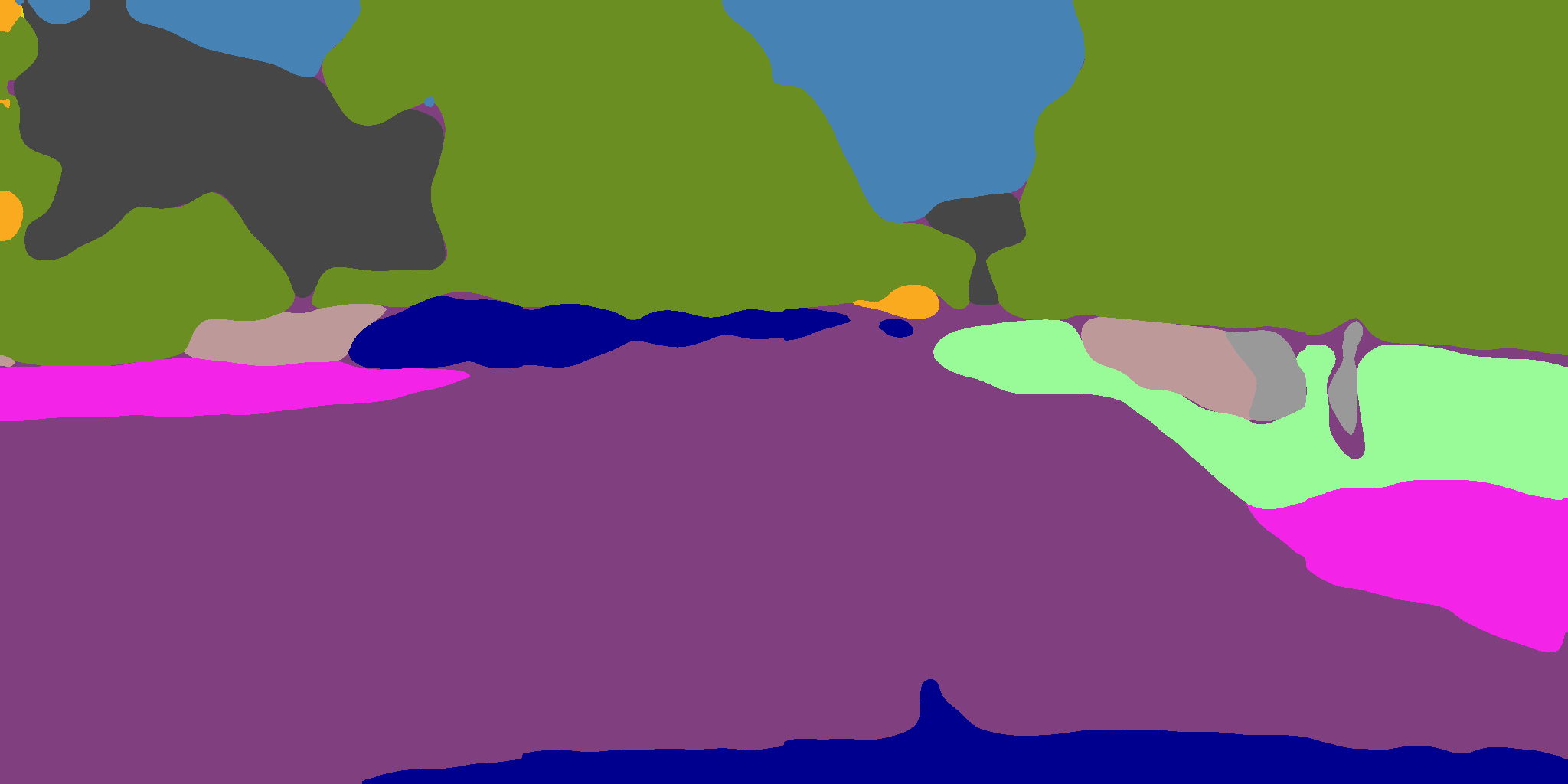}};
            \node[rectangle,draw=white, line width=0.2mm, minimum height = 1cm] at (2.7,1.05) {};
        \end{tikzpicture}
        &
        \begin{tikzpicture}
            \node[anchor=south west,inner sep=0] (image) at (0,0) {\includegraphics[width=0.18\textwidth]{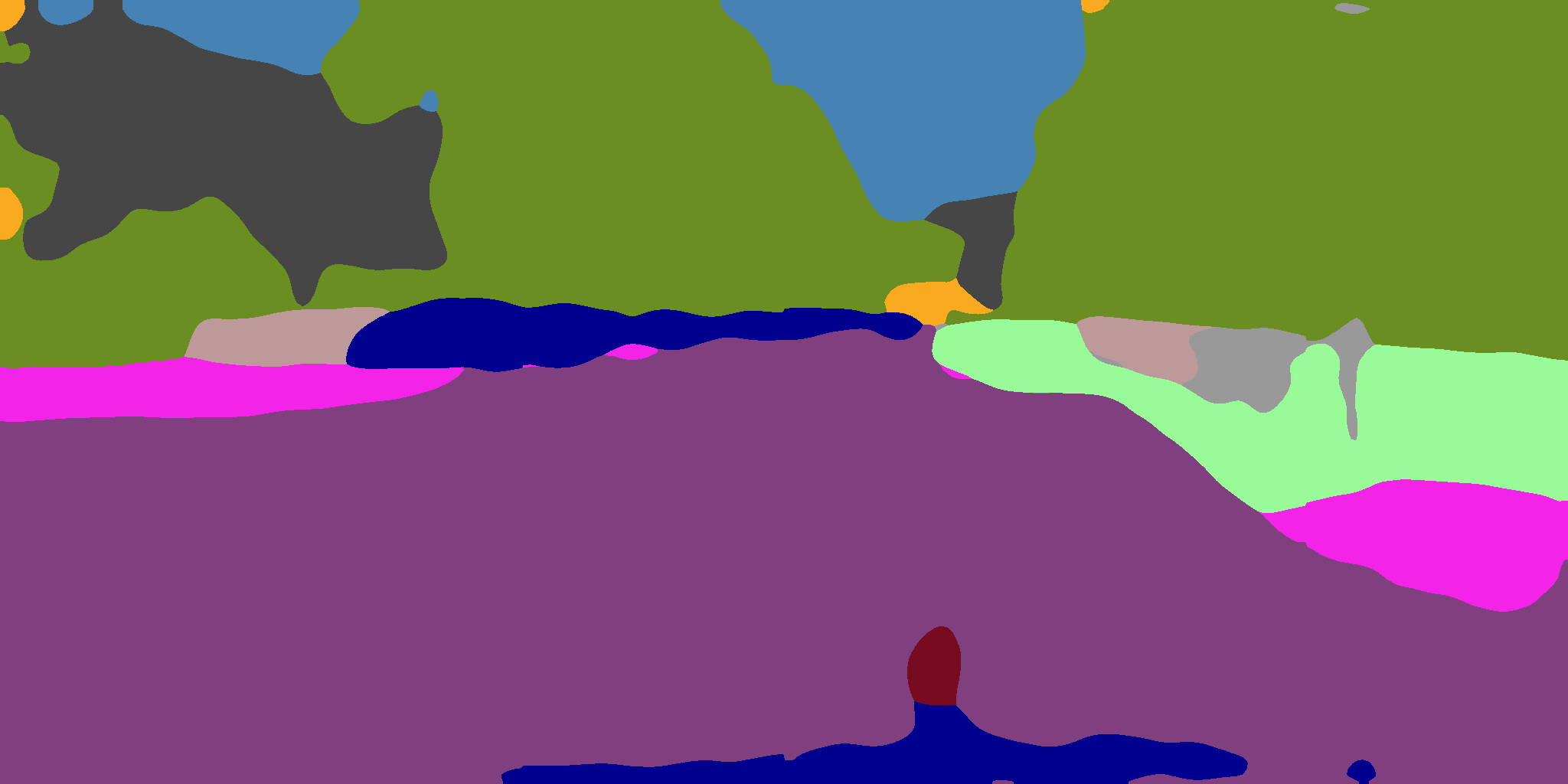}};
            \node[rectangle,draw=white, line width=0.2mm, minimum height = 1cm] at (2.7,1.05) {};
        \end{tikzpicture}
        &
        \begin{tikzpicture}
            \node[anchor=south west,inner sep=0] (image) at (0,0) {\includegraphics[width=0.18\textwidth]{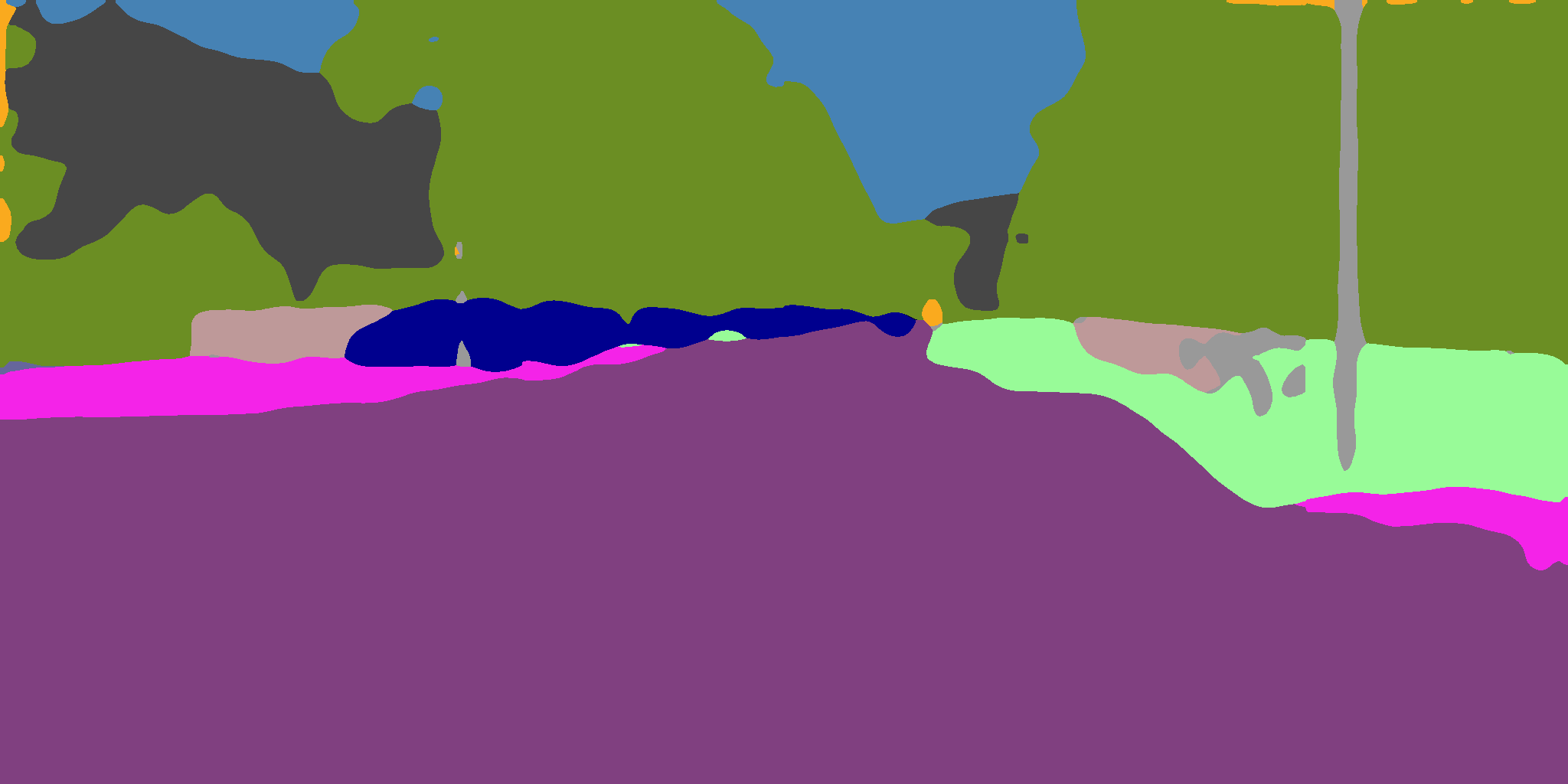}};
            \node[rectangle,draw=white, line width=0.2mm, minimum height = 1cm] at (2.7,1.05) {};
        \end{tikzpicture}
        &
        \begin{tikzpicture}
            \node[anchor=south west,inner sep=0] (image) at (0,0) {\includegraphics[width=0.18\textwidth]{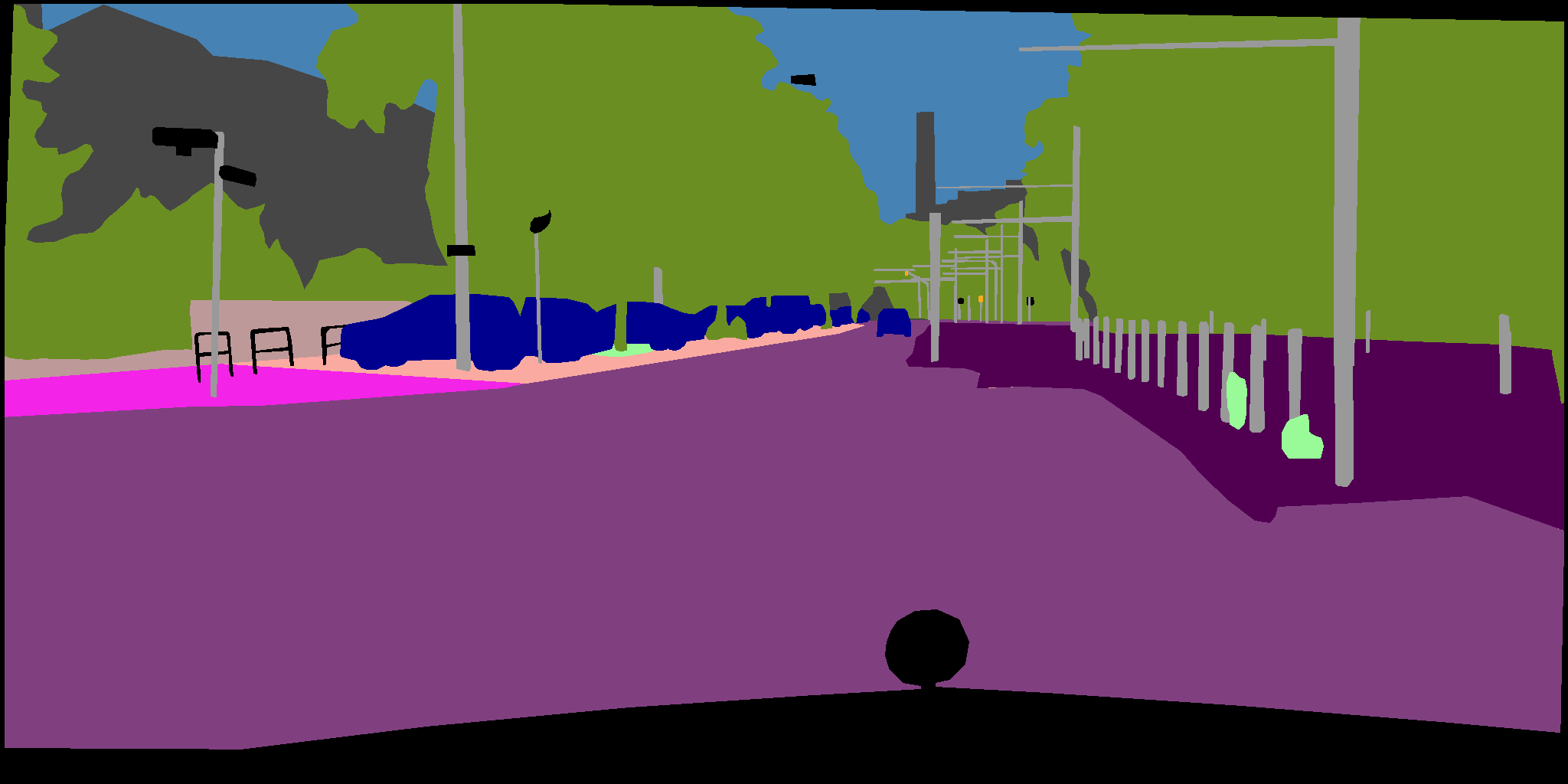}};
            \node[rectangle,draw=white, line width=0.2mm, minimum height = 1cm] at (2.7,1.05) {};
        \end{tikzpicture}\\

        \begin{tikzpicture}
            \node[anchor=south west,inner sep=0] (image) at (0,0) {\includegraphics[width=0.18\textwidth]{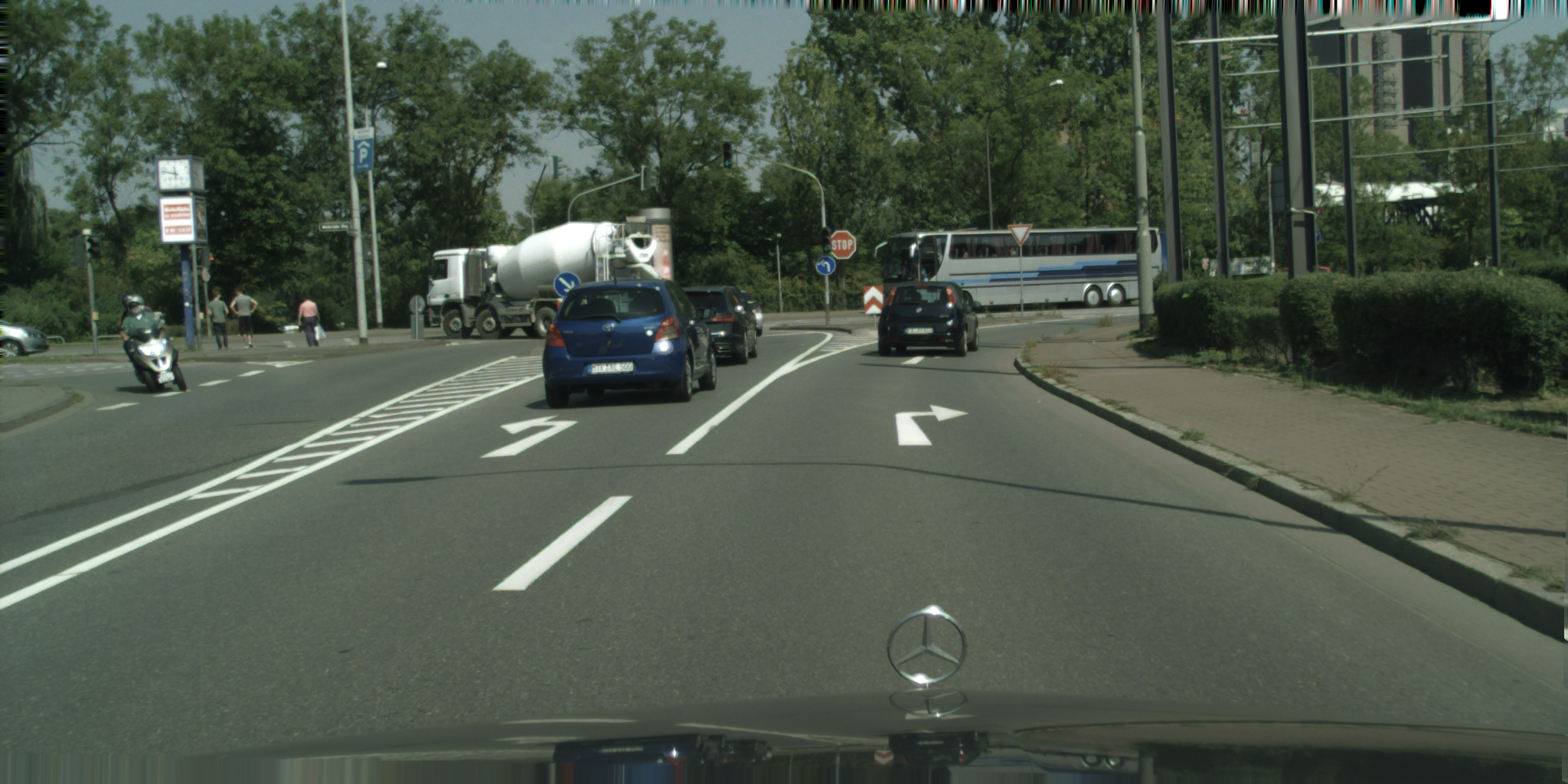}};
        \end{tikzpicture}  
        &
        \begin{tikzpicture}
            \node[anchor=south west,inner sep=0] (image) at (0,0) {\includegraphics[width=0.18\textwidth]{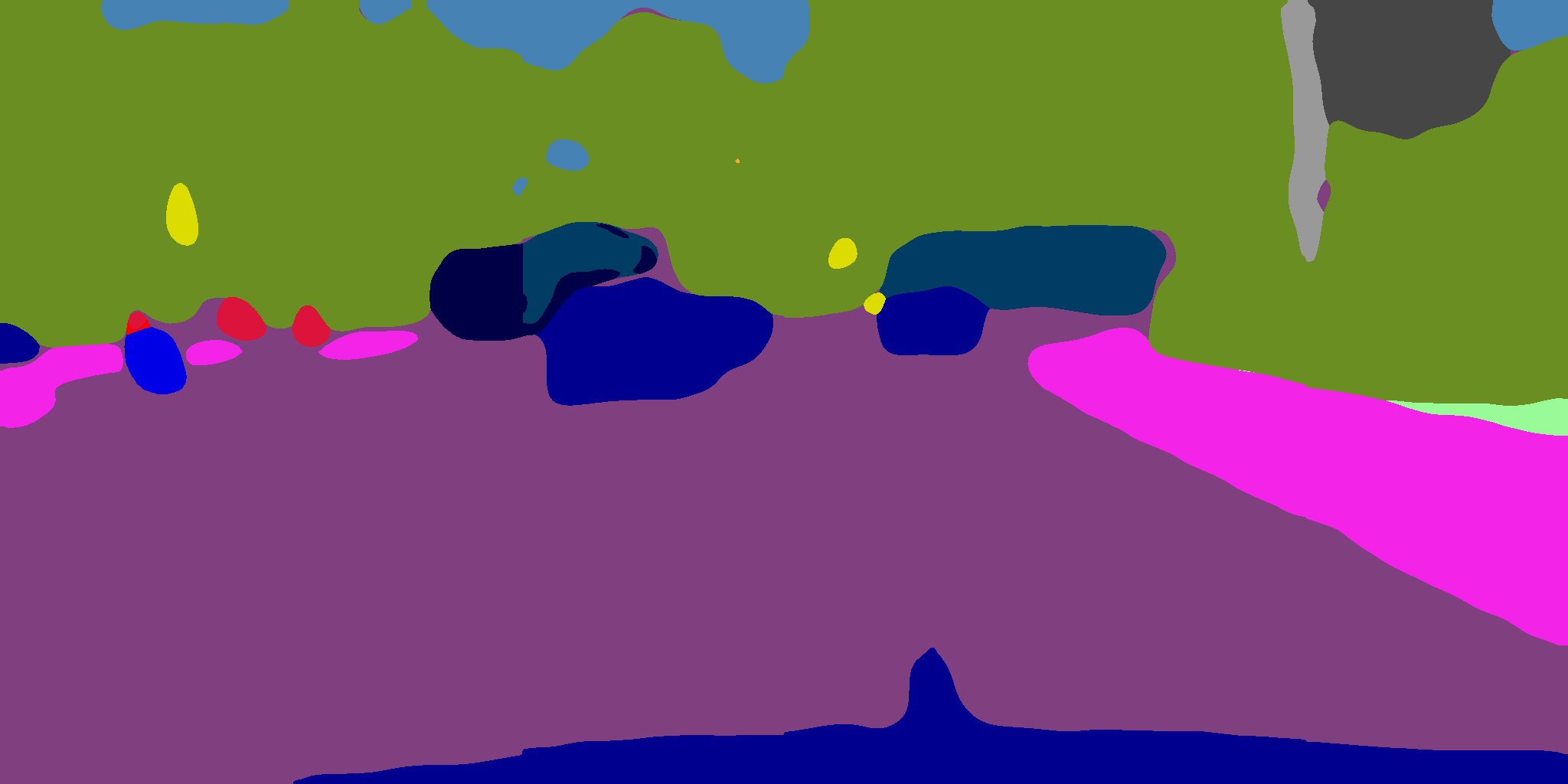}};
            \node[rectangle,draw=white, line width=0.2mm, minimum width = 0.8cm, minimum height = 0.4cm] at (.5,1) {};
        \end{tikzpicture}
        &
        \begin{tikzpicture}
            \node[anchor=south west,inner sep=0] (image) at (0,0) {\includegraphics[width=0.18\textwidth]{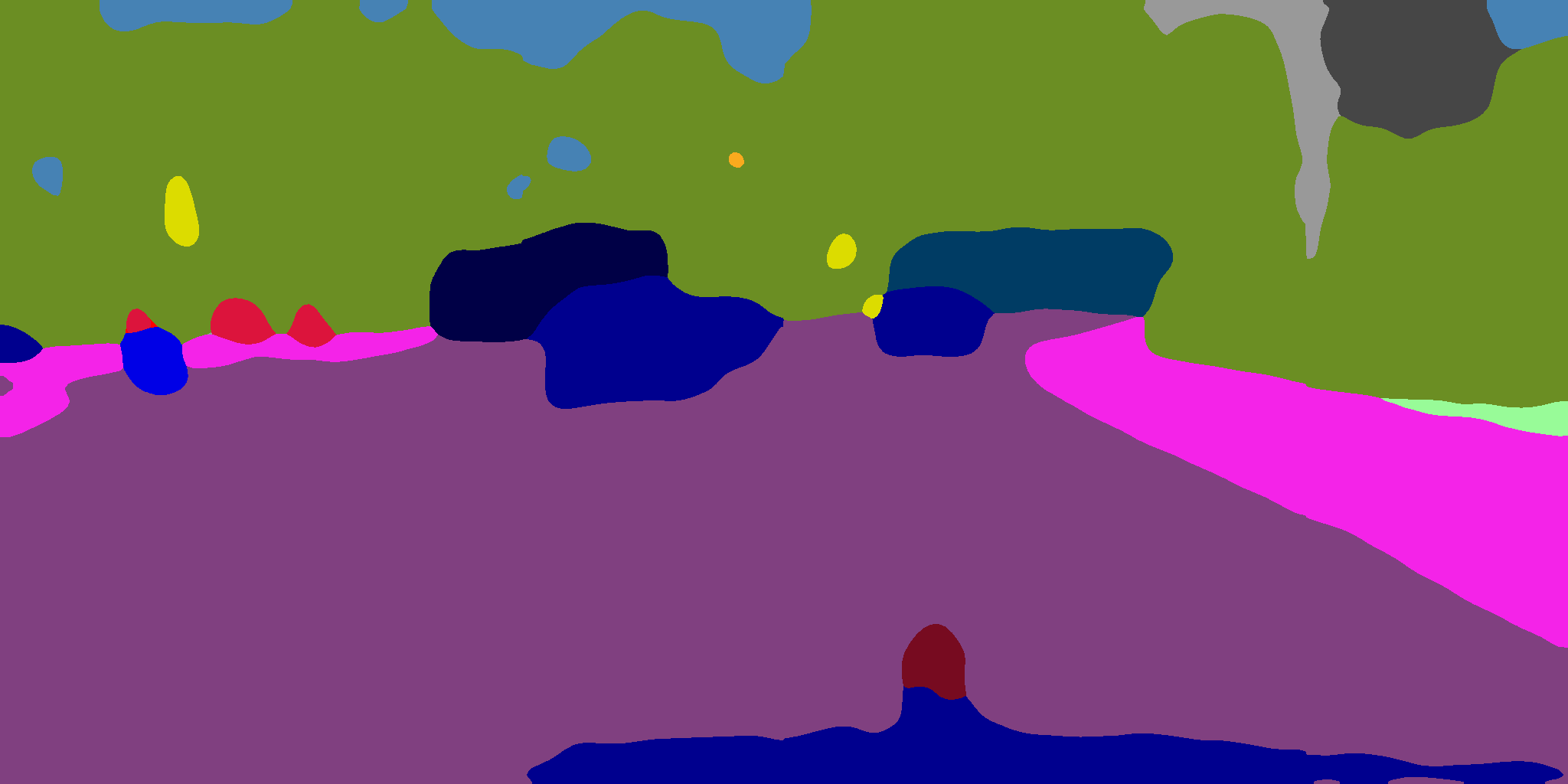}};
            \node[rectangle,draw=white, line width=0.2mm, minimum width = 0.8cm, minimum height = 0.4cm] at (.5,1) {};
        \end{tikzpicture}
        &
        \begin{tikzpicture}
            \node[anchor=south west,inner sep=0] (image) at (0,0) {\includegraphics[width=0.18\textwidth]{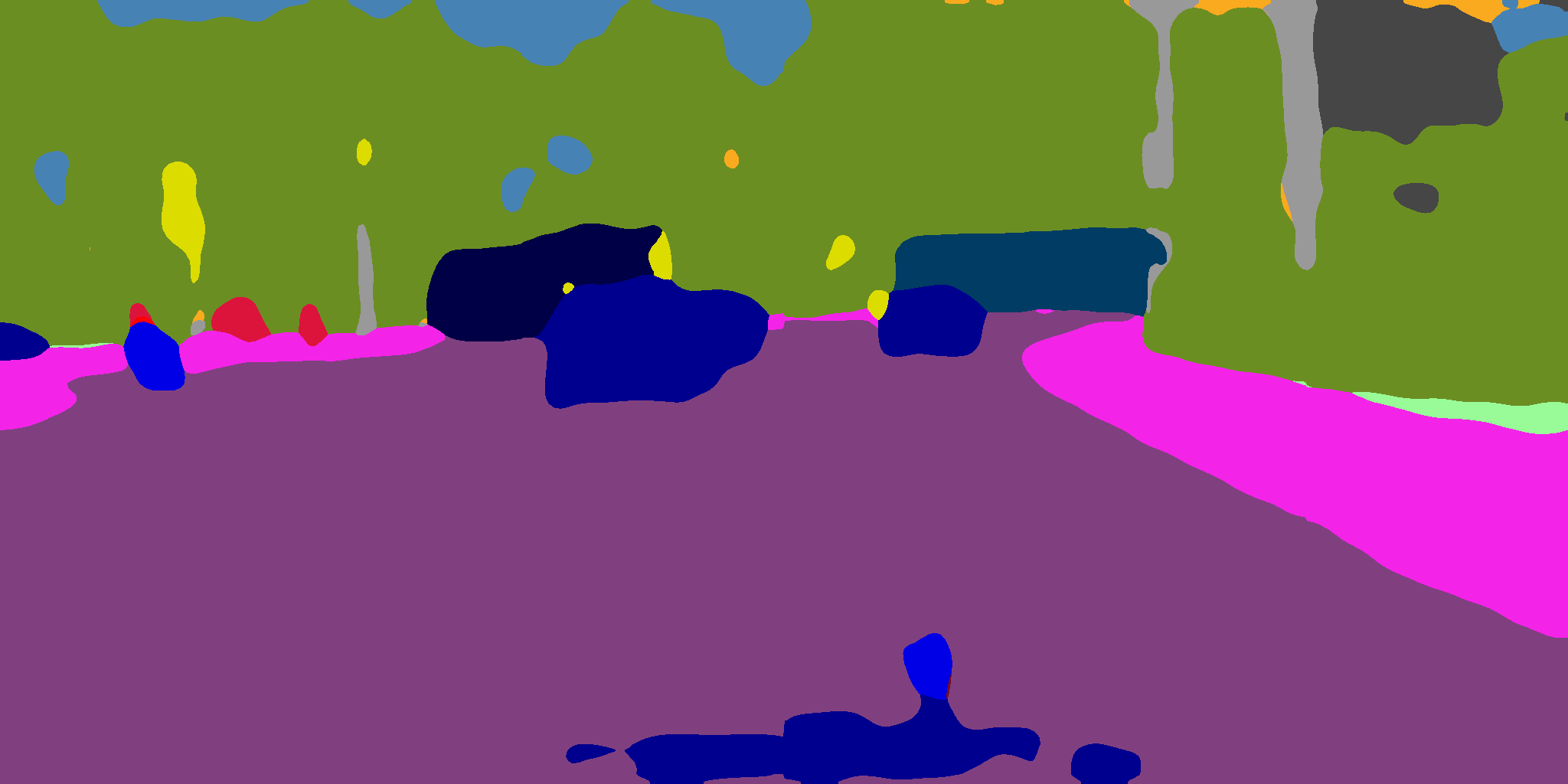}};
            \node[rectangle,draw=white, line width=0.2mm, minimum width = 0.8cm, minimum height = 0.4cm] at (.5,1) {};
        \end{tikzpicture}
        &
        \begin{tikzpicture}
            \node[anchor=south west,inner sep=0] (image) at (0,0) {\includegraphics[width=0.18\textwidth]{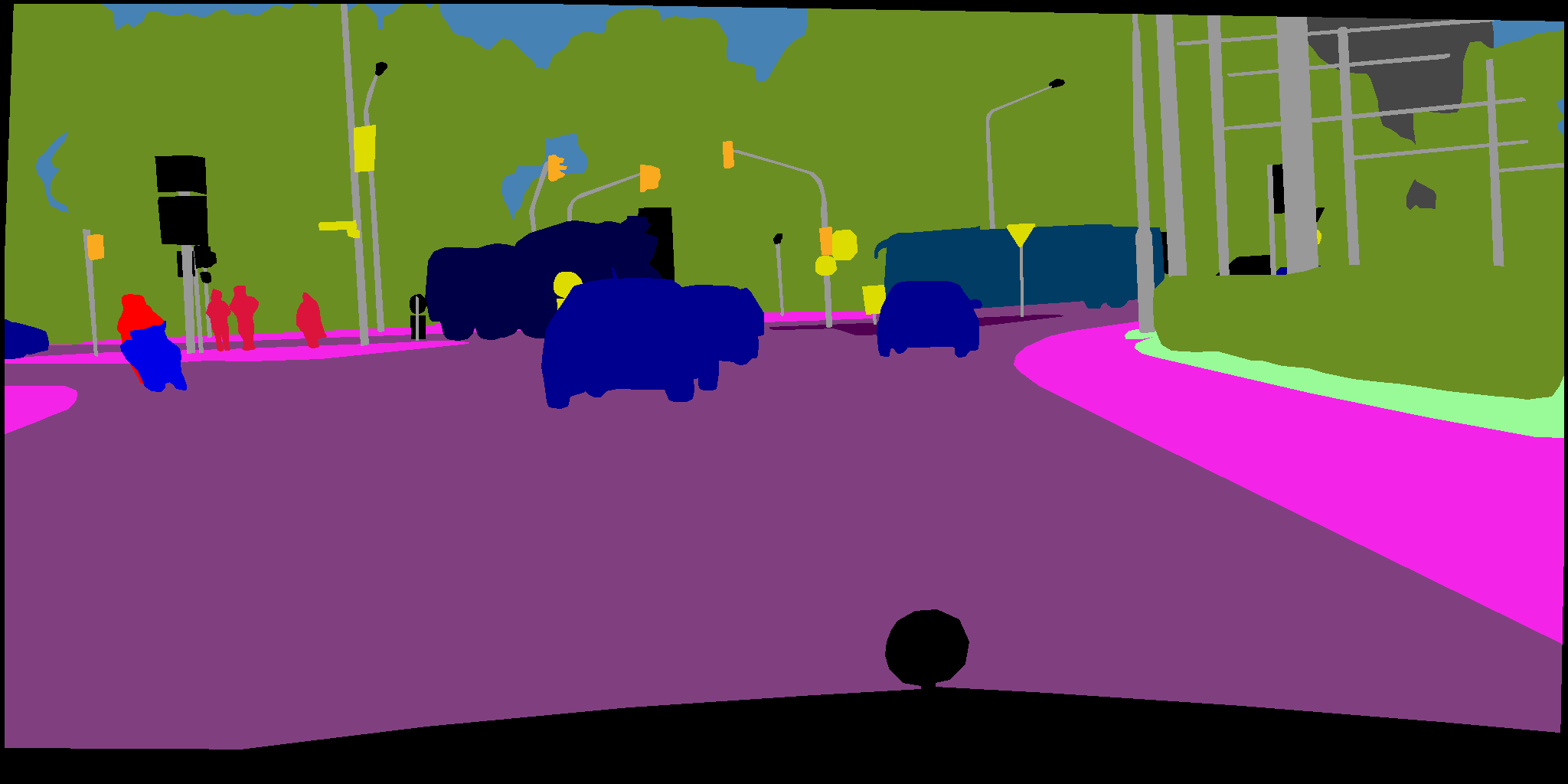}};
            \node[rectangle,draw=white, line width=0.2mm, minimum width = 0.8cm, minimum height = 0.4cm] at (.5,1) {};
        \end{tikzpicture}\\

        \begin{tikzpicture}
            \node[anchor=south west,inner sep=0] (image) at (0,0) {\includegraphics[width=0.18\textwidth]{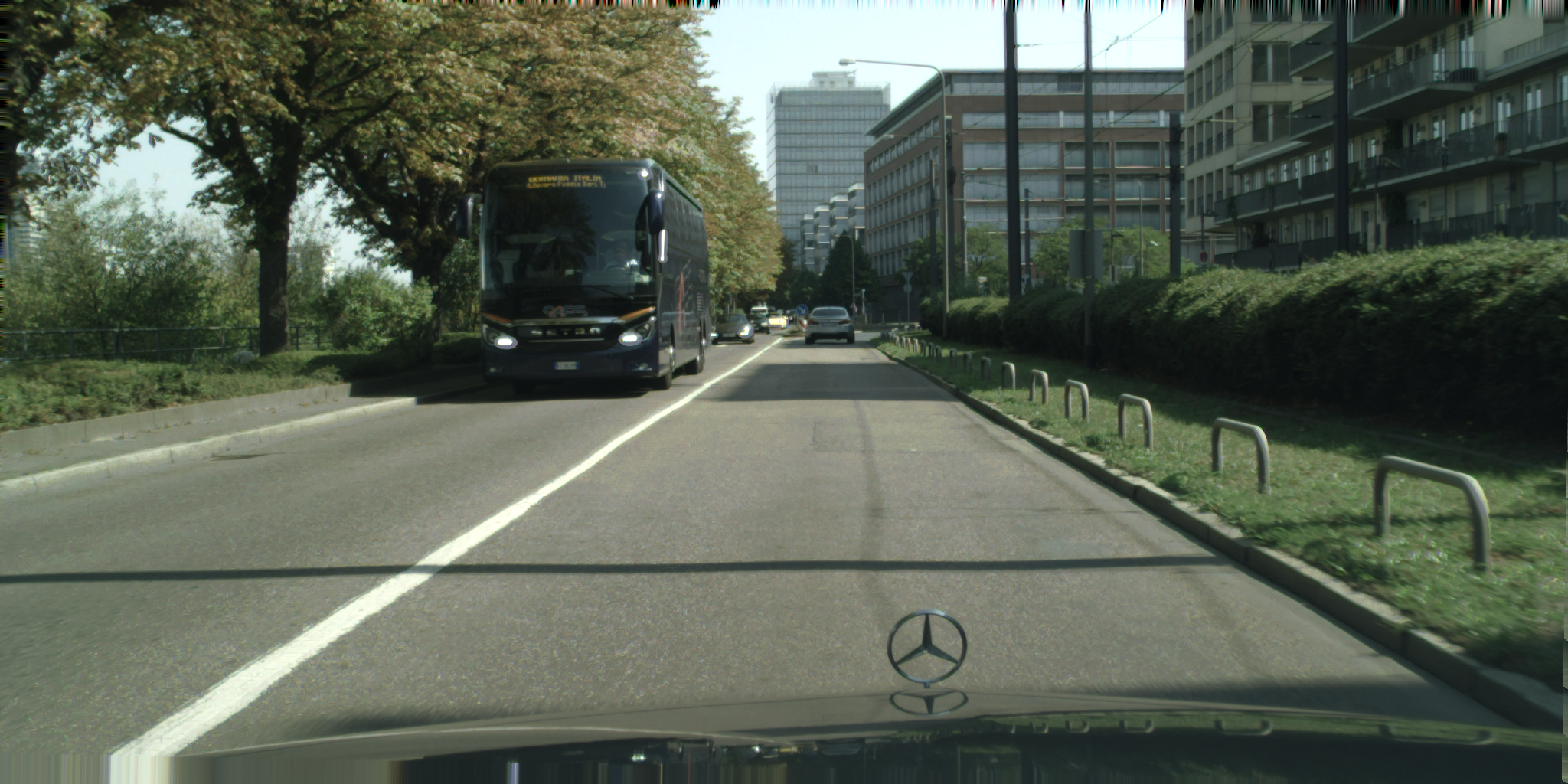}};
        \end{tikzpicture}  
        &
        \begin{tikzpicture}
            \node[anchor=south west,inner sep=0] (image) at (0,0) {\includegraphics[width=0.18\textwidth]{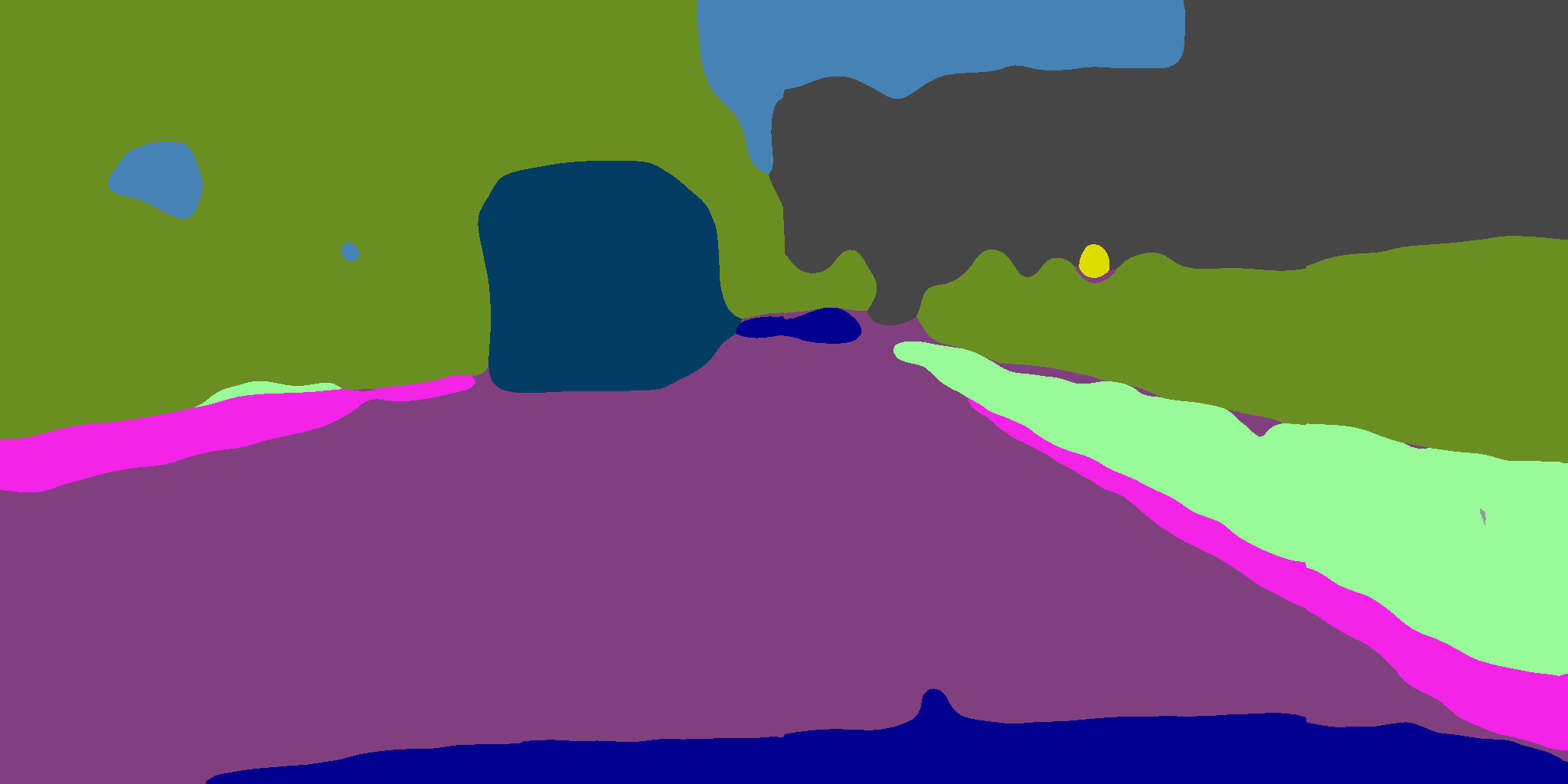}};
            \node[rectangle,draw=white, line width=0.2mm, minimum width = 1.cm, minimum height = 0.5cm] at (2.5,0.7) {};
        \end{tikzpicture}
        &
        \begin{tikzpicture}
            \node[anchor=south west,inner sep=0] (image) at (0,0) {\includegraphics[width=0.18\textwidth]{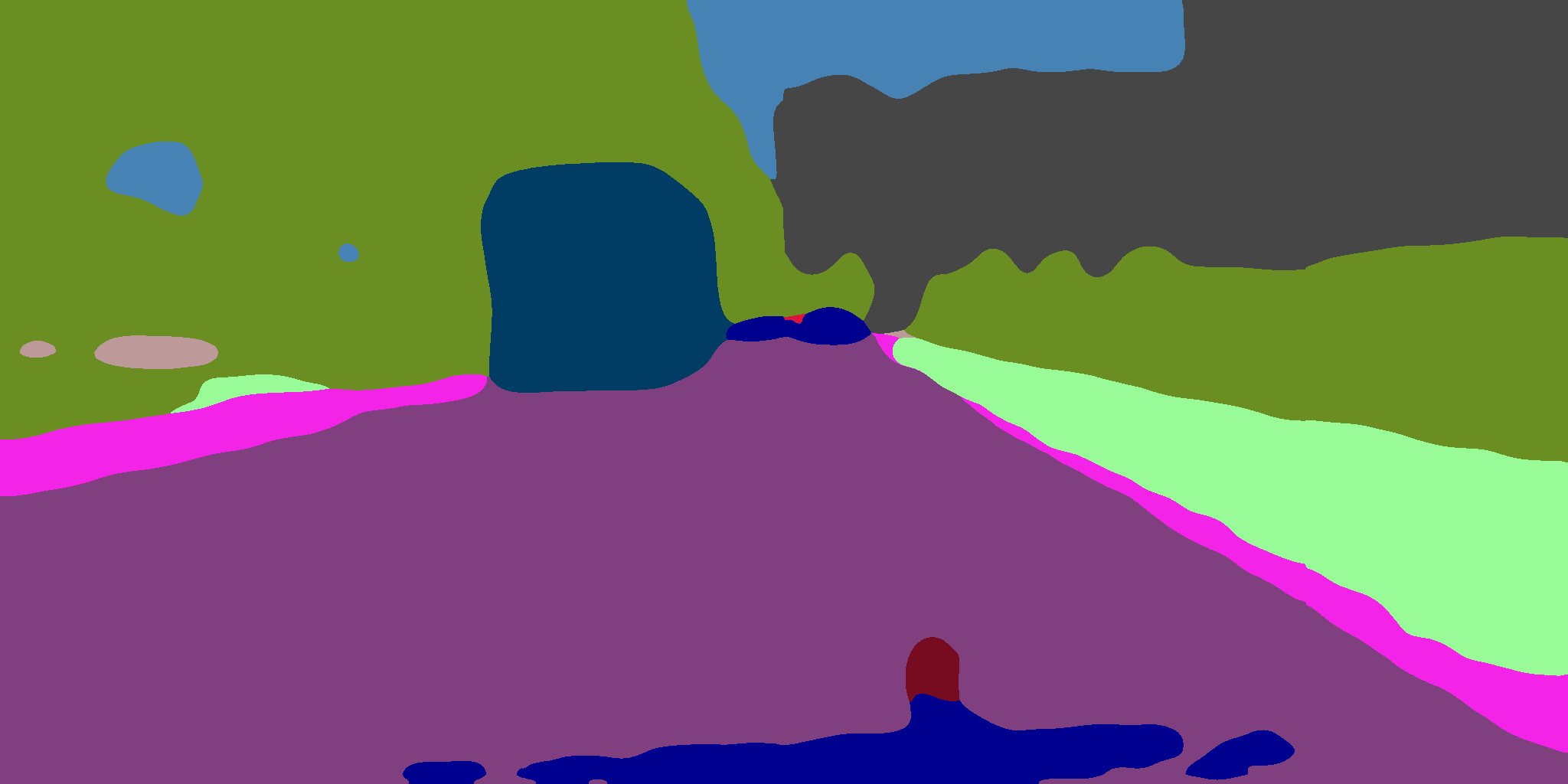}};
            \node[rectangle,draw=white, line width=0.2mm, minimum width = 1.cm, minimum height = 0.5cm] at (2.5,0.7) {};
        \end{tikzpicture}
        &
        \begin{tikzpicture}
            \node[anchor=south west,inner sep=0] (image) at (0,0) {\includegraphics[width=0.18\textwidth]{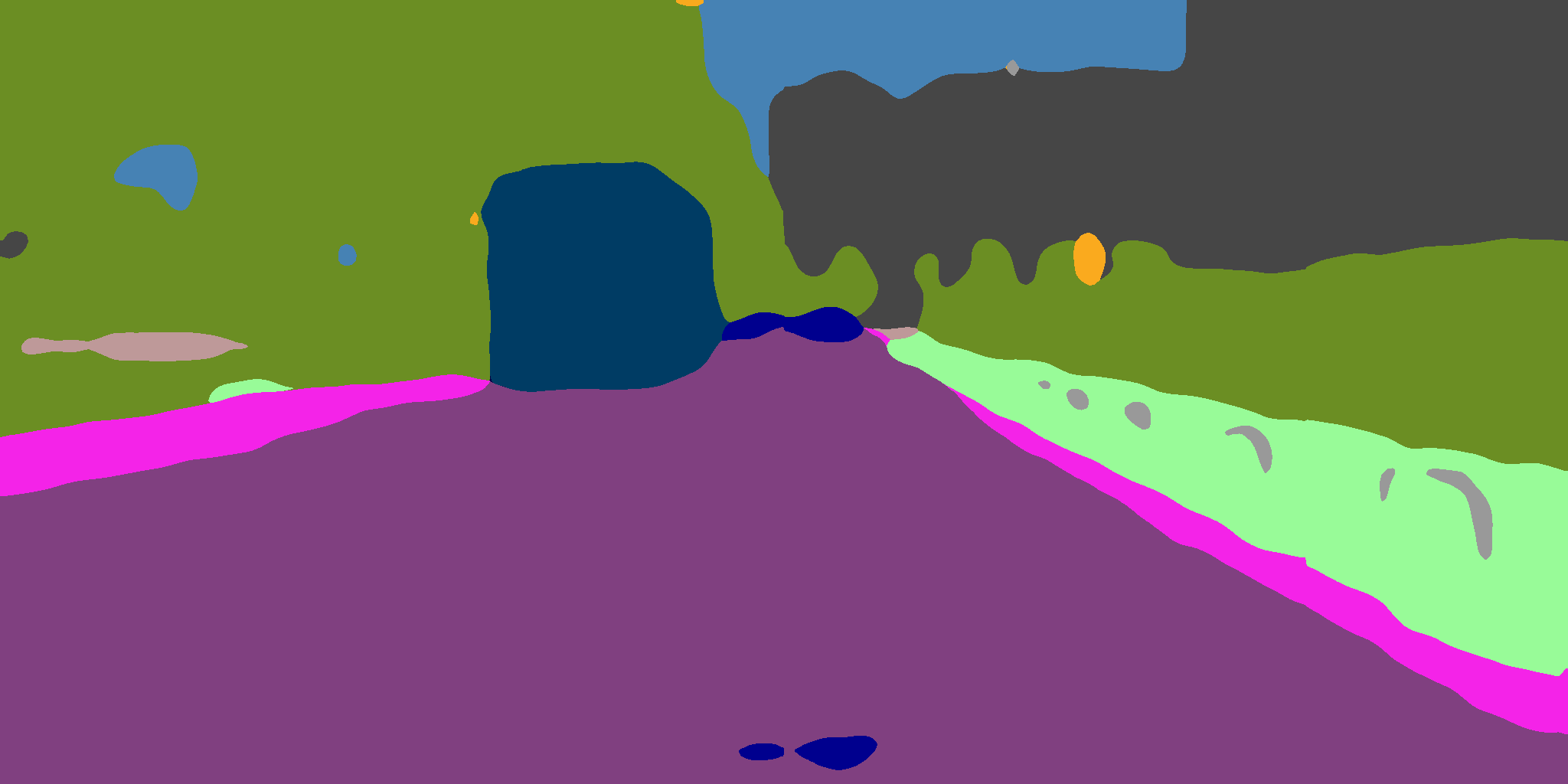}};
            \node[rectangle,draw=white, line width=0.2mm, minimum width = 1.cm, minimum height = 0.5cm] at (2.5,0.7) {};
        \end{tikzpicture}
        &
        \begin{tikzpicture}
            \node[anchor=south west,inner sep=0] (image) at (0,0) {\includegraphics[width=0.18\textwidth]{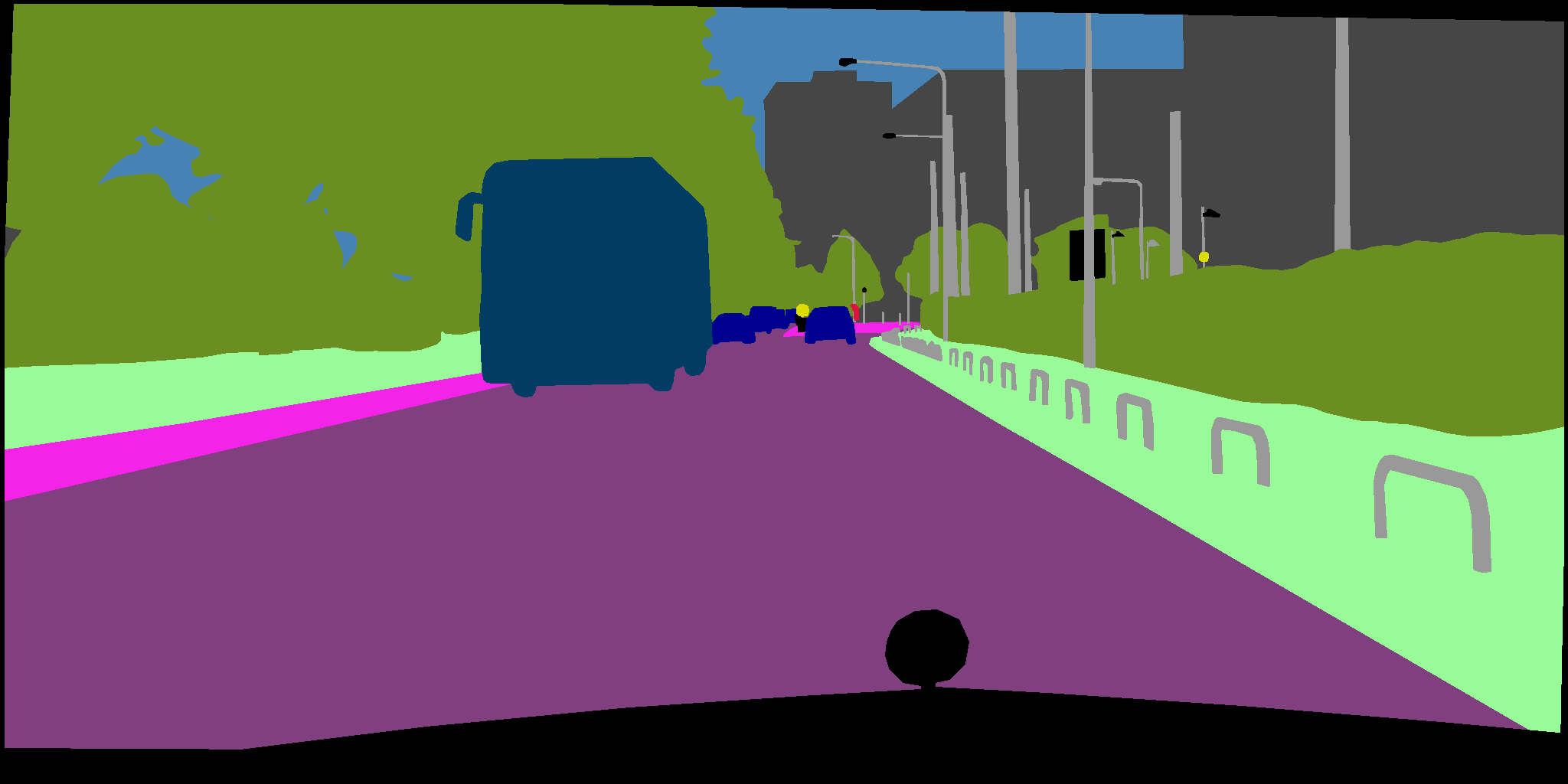}};
            \node[rectangle,draw=white, line width=0.2mm, minimum width = 1.cm, minimum height = 0.5cm] at (2.5,0.7) {};
        \end{tikzpicture}\\

        \begin{tikzpicture}
            \node[anchor=south west,inner sep=0] (image) at (0,0) {\includegraphics[width=0.18\textwidth]{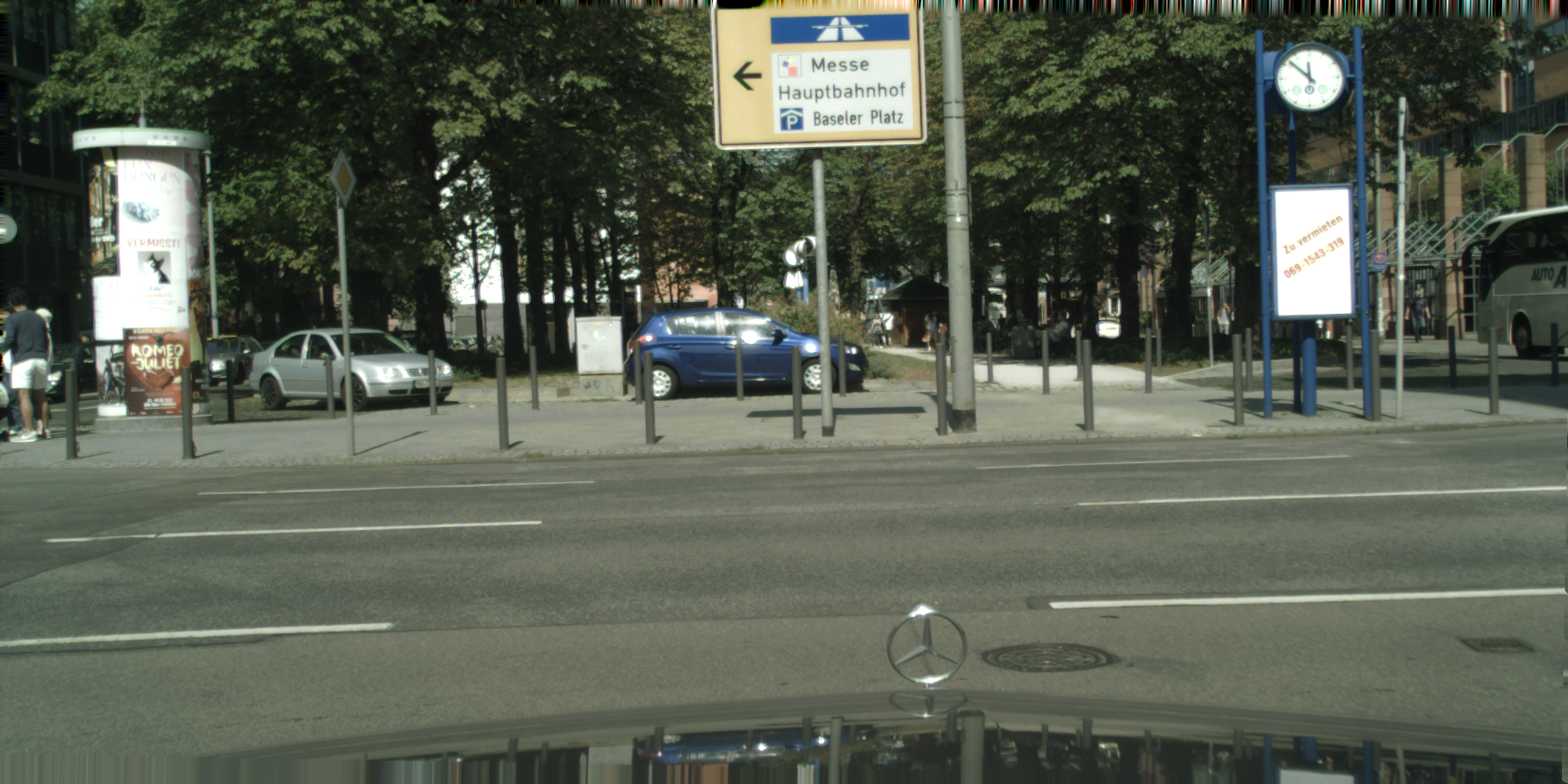}};
        \end{tikzpicture}  
        &
        \begin{tikzpicture}
            \node[anchor=south west,inner sep=0] (image) at (0,0) {\includegraphics[width=0.18\textwidth]{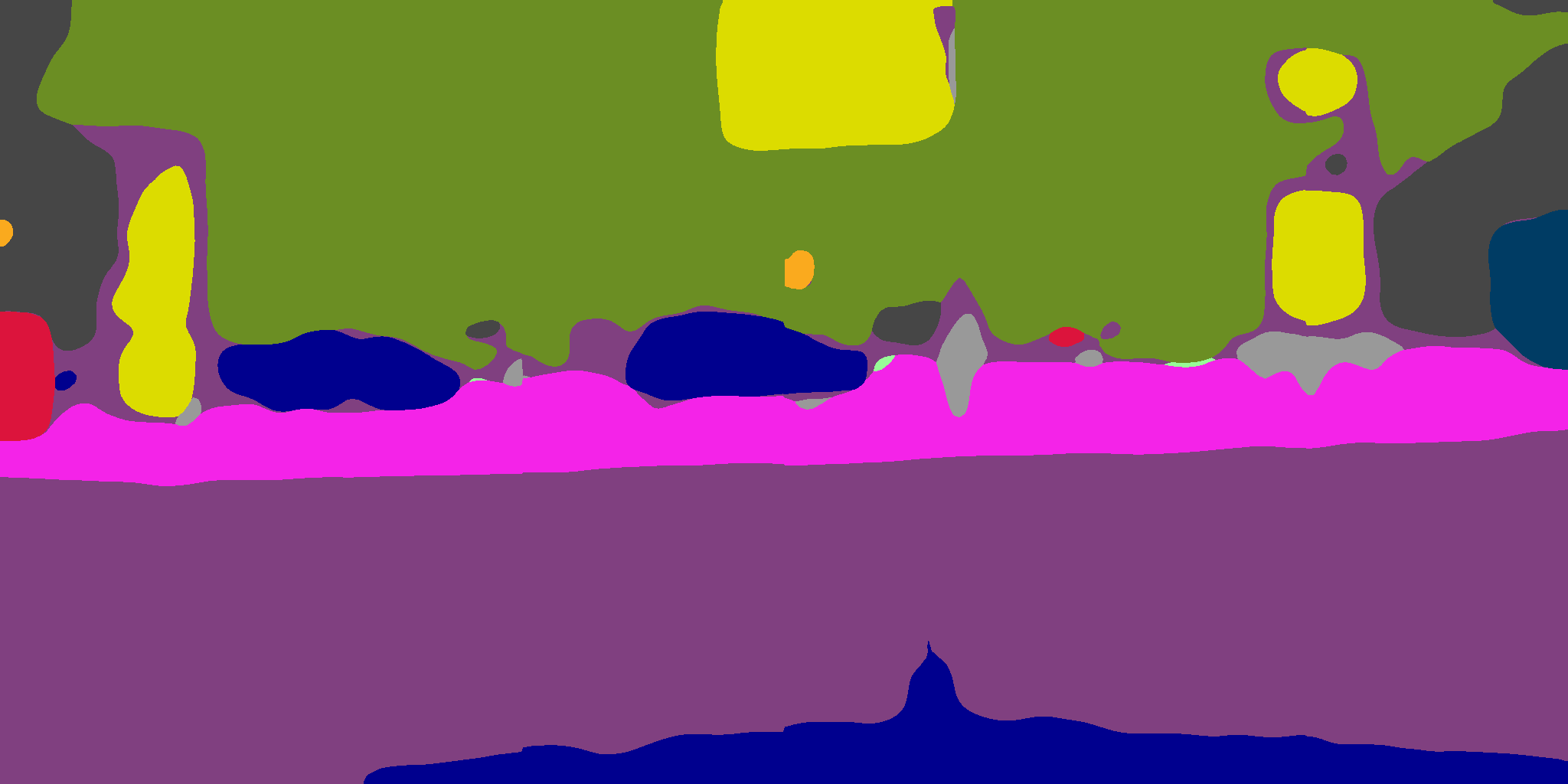}};
            \node[rectangle,draw=white, line width=0.2mm, minimum width = .3cm, minimum height = 0.9cm] at (1.9,1.1) {};
        \end{tikzpicture}
        &
        \begin{tikzpicture}
            \node[anchor=south west,inner sep=0] (image) at (0,0) {\includegraphics[width=0.18\textwidth]{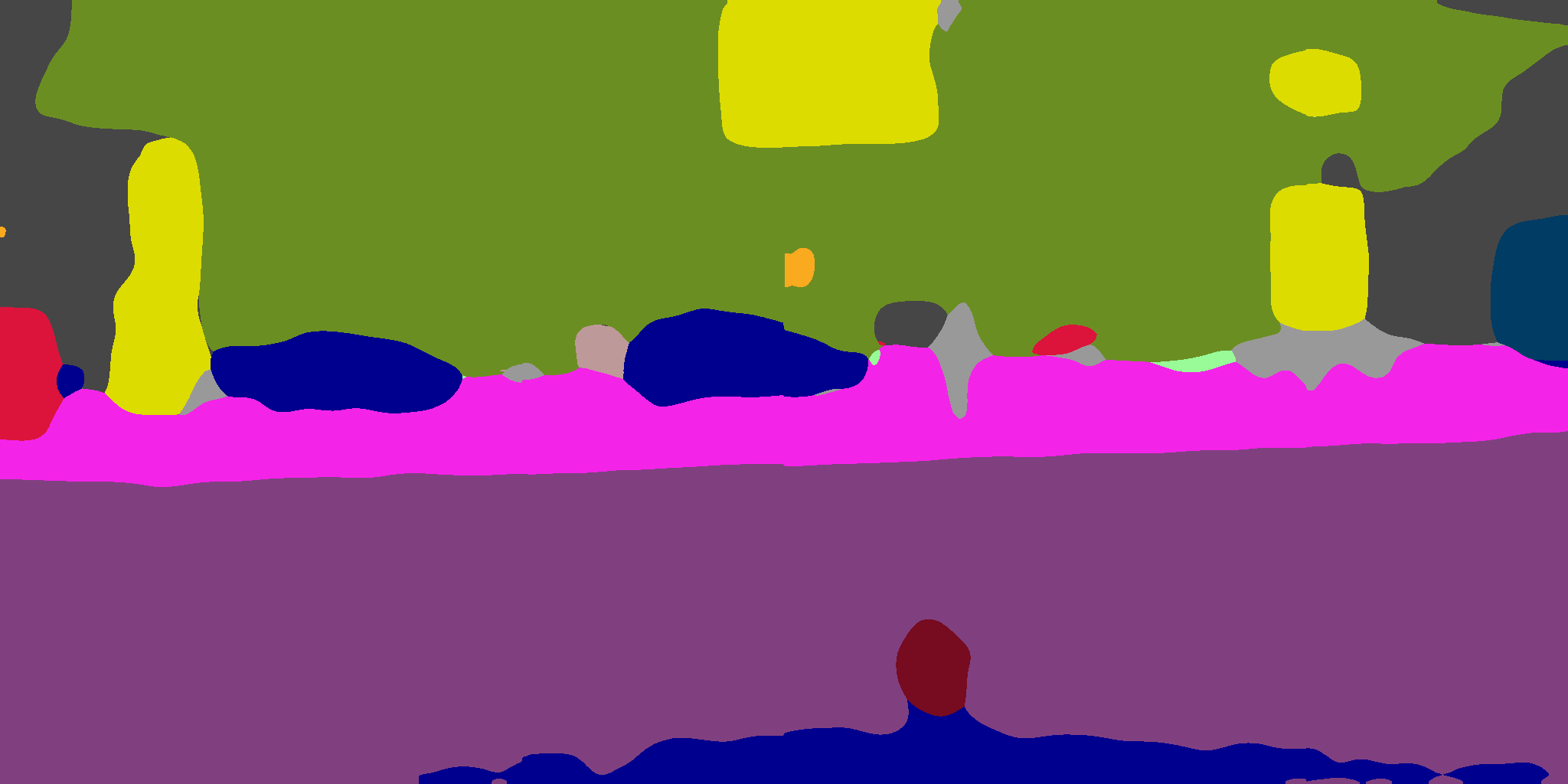}};
            \node[rectangle,draw=white, line width=0.2mm, minimum width = .3cm, minimum height = 0.9cm] at (1.9,1.1) {};
        \end{tikzpicture}
        &
        \begin{tikzpicture}
            \node[anchor=south west,inner sep=0] (image) at (0,0) {\includegraphics[width=0.18\textwidth]{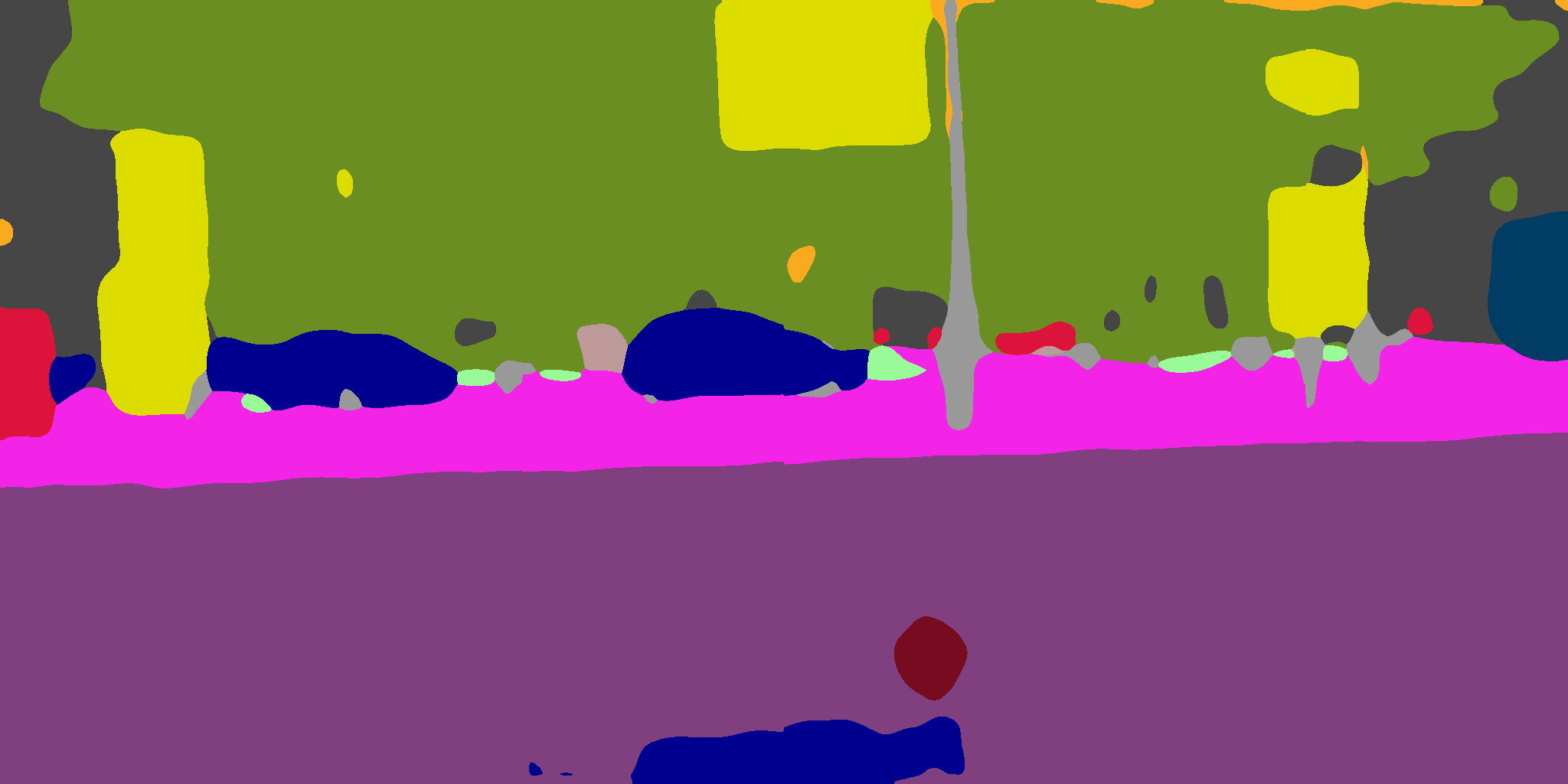}};
            \node[rectangle,draw=white, line width=0.2mm, minimum width = .3cm, minimum height = 0.9cm] at (1.9,1.1) {};
        \end{tikzpicture}
        &
        \begin{tikzpicture}
            \node[anchor=south west,inner sep=0] (image) at (0,0) {\includegraphics[width=0.18\textwidth]{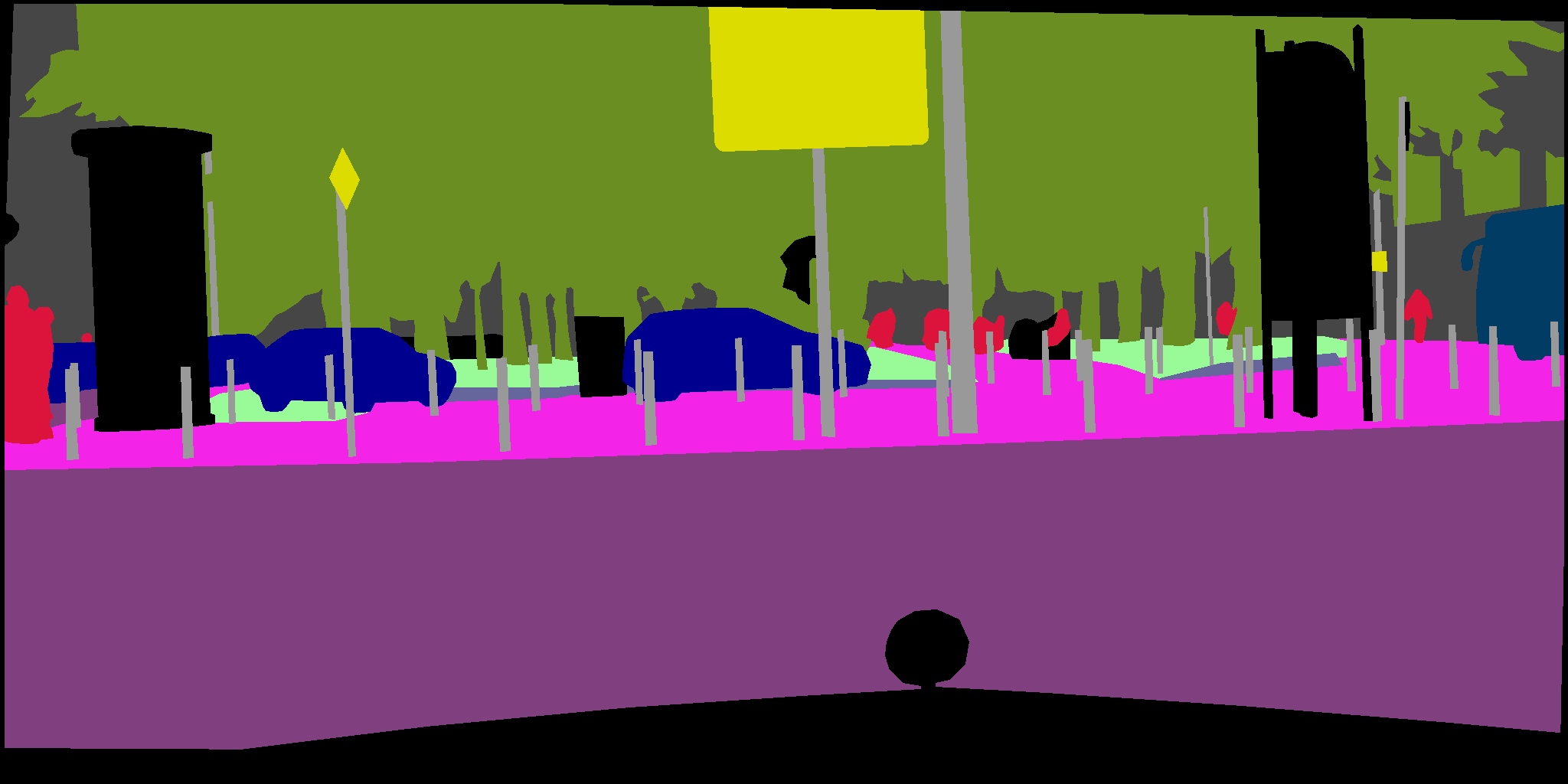}};
            \node[rectangle,draw=white, line width=0.2mm, minimum width = .3cm, minimum height = 0.9cm] at (1.9,1.1) {};
        \end{tikzpicture}\\
        
        Input & CoRTe~\cite{cuttano2023cross} & Naive Transfer & \method (Ours) & Ground truth \\
    \end{tabular}
    \captionof{figure}{\textbf{Qualitative comparison of different methods.} From left to right: input RGB image, predictions from CoRTE~\cite{cuttano2023cross}, Naive Transfer, our method \method (Ours), and ground-truth segmentation maps. Our method shows improved segmentation quality of small objects, such as ``poles'' (highlighted with white rectangular boxes), compared to the other approaches.}
    \label{fig:qualitative_comparison}
\end{table*}

%% file: Tables/resolution_supp.tex
\begin{figure*}
    \centering
    \includegraphics[clip, trim=0.0cm 0.cm 15.5cm 0.0cm, width=0.48\textwidth]{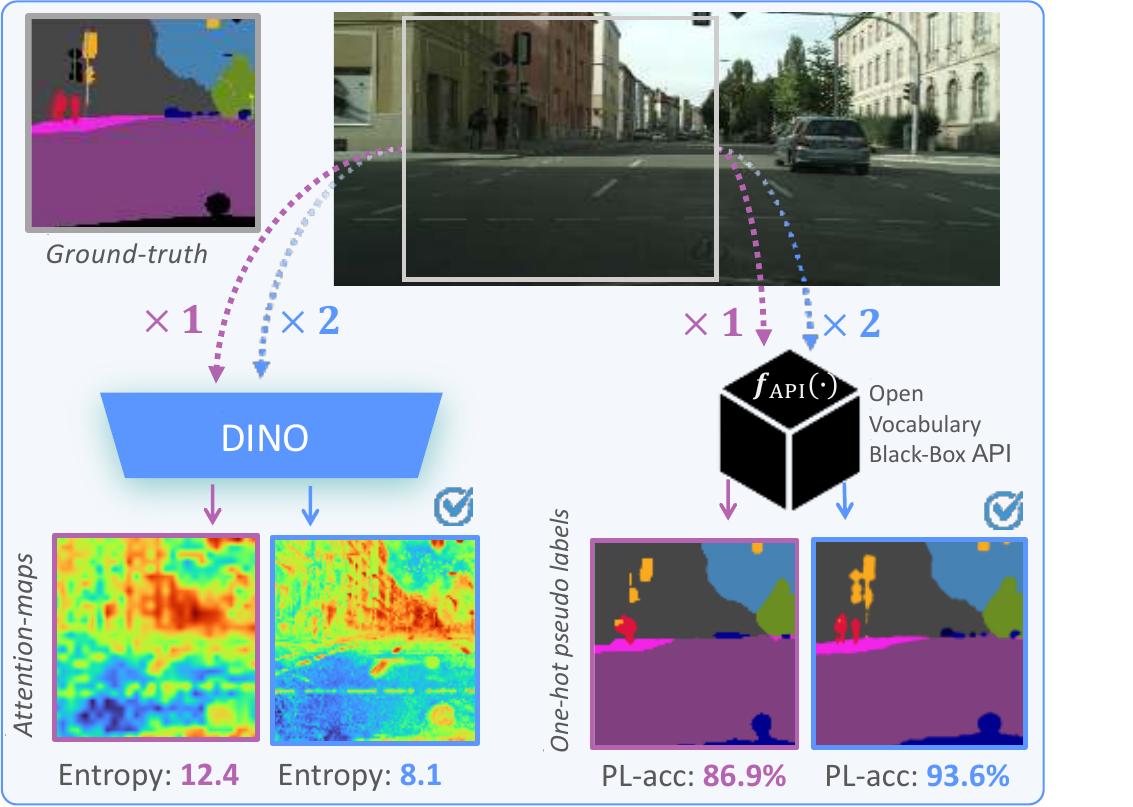}
    \includegraphics[clip, trim=0.0cm 0.cm 15.5cm 0.0cm, width=0.48\textwidth]{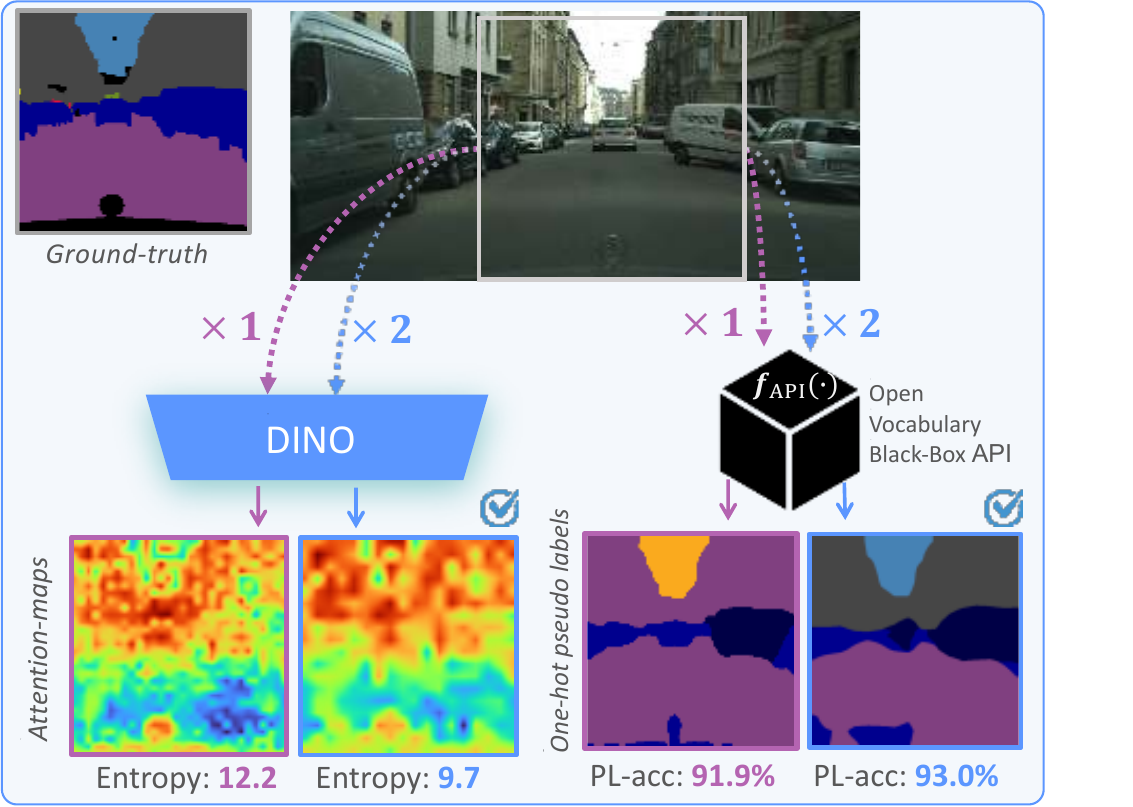}
       \caption{\textbf{DINOv2 and [\texttt{CLS}] Token for Optimal Resolution Selection.} DINOv2's [\texttt{CLS}] token attention maps are computed at multiple scales for each input image. The resolution with the highest spatially averaged attention score (\emph{e.g.}, $\times1$, $\times 1.5$ and $\times2$) is selected to generate pseudo-labels from the API model.}
    \label{fig:affinity-segmaps}
\end{figure*}

%% file: Tables/failure_supp.tex
\begin{figure*}[t]
    \centering

    \begin{minipage}[b]{0.48\textwidth}
        \centering
        \includegraphics[width=0.32\textwidth]{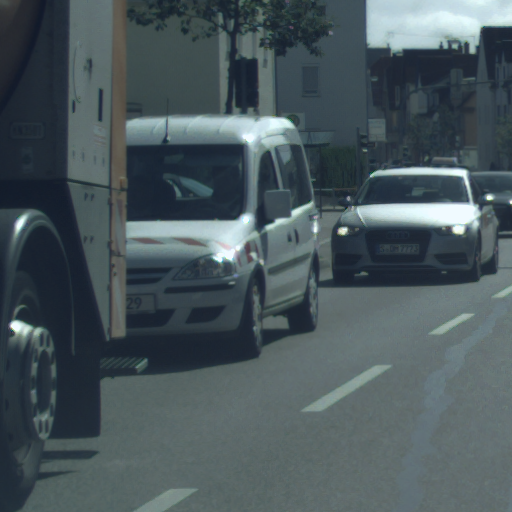}
        \hfill
        \includegraphics[width=0.32\textwidth]{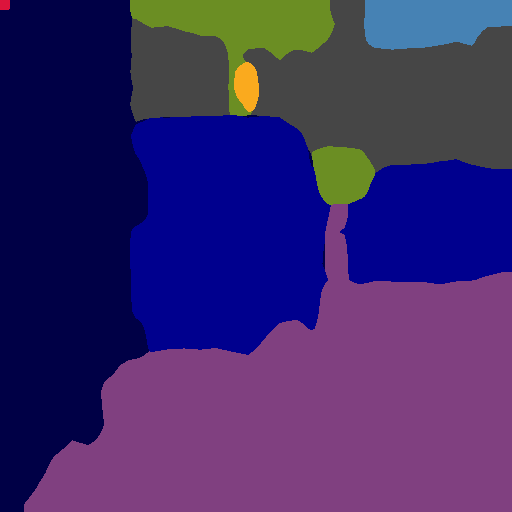}
        \hfill
        \includegraphics[width=0.32\textwidth]{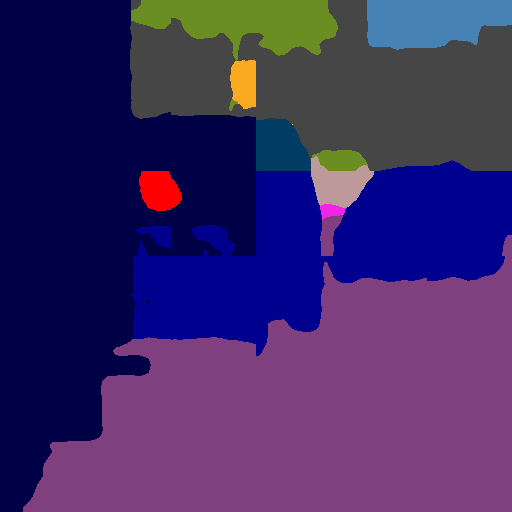}
        \vspace{0.3em}

        {\small Example 1: (left) RGB input, (middle) prediction on $\Xmat_c$, (right) prediction on $\Xmat_c^*$}
    \end{minipage}
    \hfill
    \begin{minipage}[b]{0.48\textwidth}
        \centering
        \includegraphics[width=0.32\textwidth]{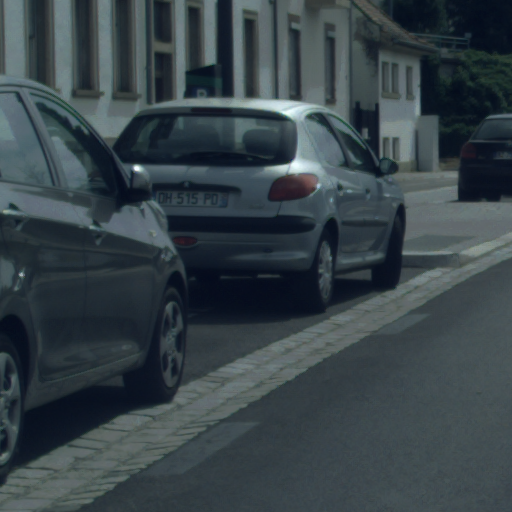}
        \hfill
        \includegraphics[width=0.32\textwidth]{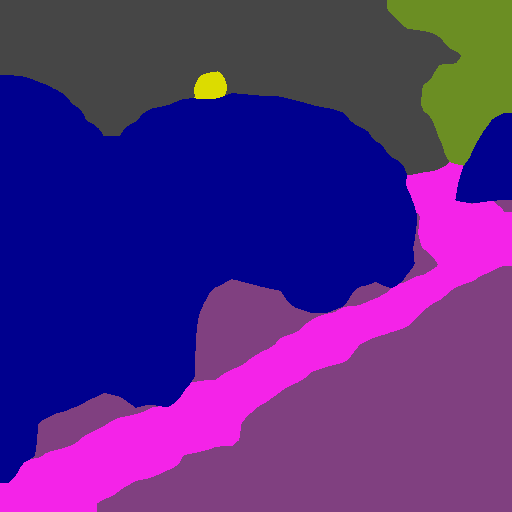}
        \hfill
        \includegraphics[width=0.32\textwidth]{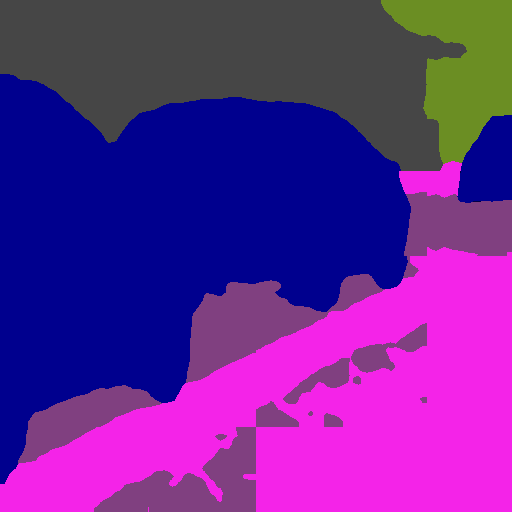}
        \vspace{0.3em}

        {\small Example 2: (left) RGB input, (middle) prediction on $\Xmat_c$, (right) prediction on $\Xmat_c^*$}
    \end{minipage}

    \vspace{1em}

    \begin{minipage}[b]{0.48\textwidth}
        \centering
        \includegraphics[width=0.32\textwidth]{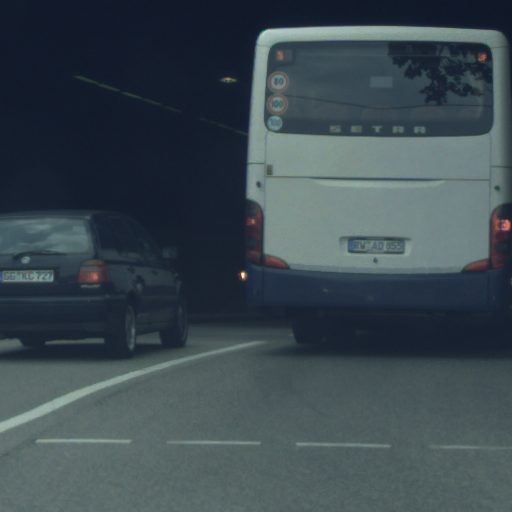}
        \hfill
        \includegraphics[width=0.32\textwidth]{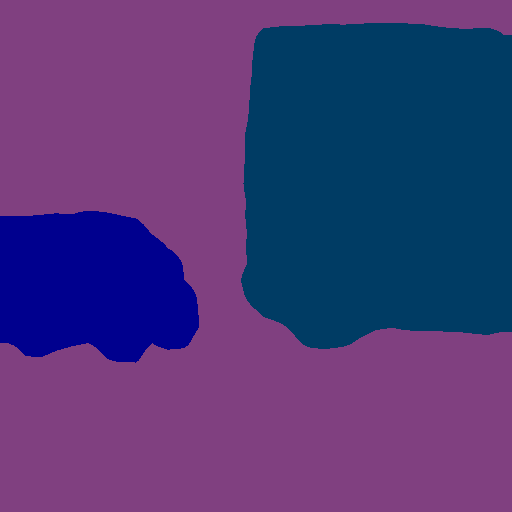}
        \hfill
        \includegraphics[width=0.32\textwidth]{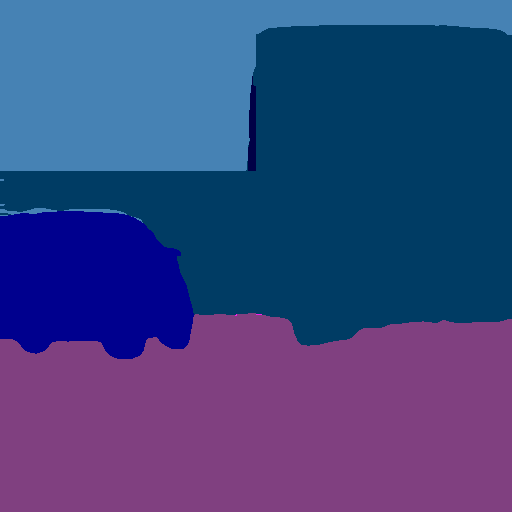}
        \vspace{0.3em}

        {\small Example 3: (left) RGB input, (middle) prediction on $\Xmat_c$, (right) prediction on $\Xmat_c^*$}
    \end{minipage}
    \hfill
    \begin{minipage}[b]{0.48\textwidth}
        \centering
        \includegraphics[width=0.32\textwidth]{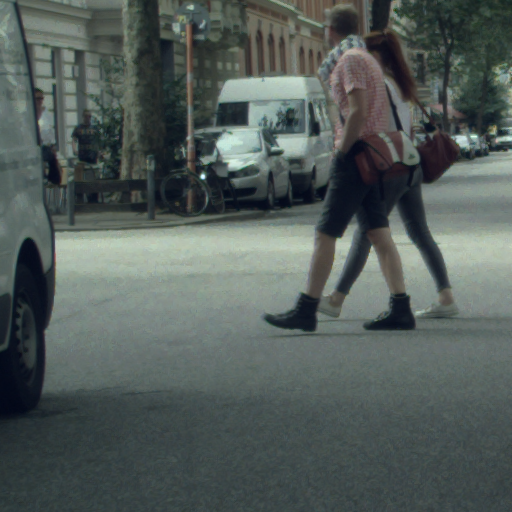}
        \hfill
        \includegraphics[width=0.32\textwidth]{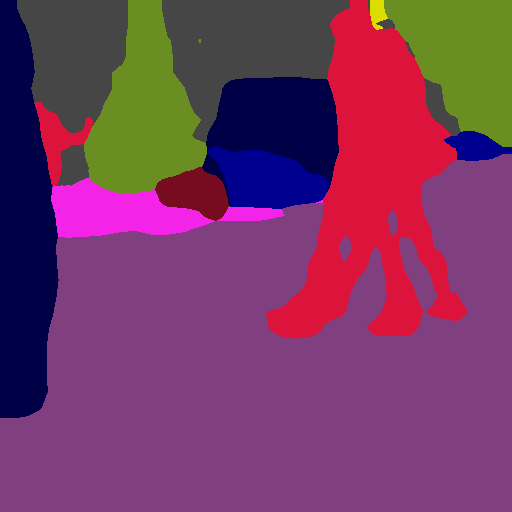}
        \hfill
        \includegraphics[width=0.32\textwidth]{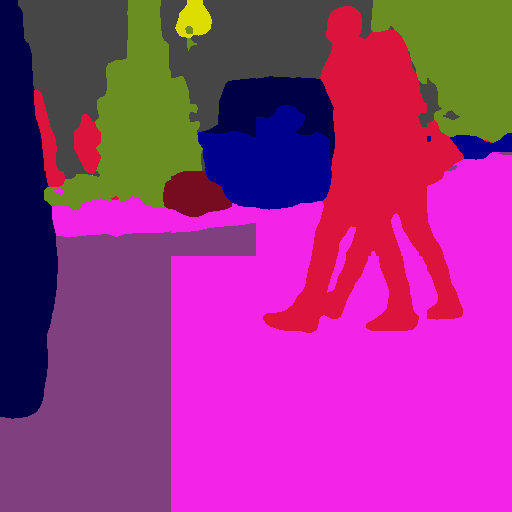}
        \vspace{0.3em}

        {\small Example 4: (left) RGB input, (middle) prediction on $\Xmat_c$, (right) prediction on $\Xmat_c^*$}
    \end{minipage}

    \caption{\textbf{Failure cases of our framework}. The figure illustrates cases where the pseudo-label given by the API model on the cropped image ($\Xmat_c$) is more accurate than the one given by the API using the optimal scale ($\Xmat_c^*$). Despite leveraging the optimal scale, the API may struggle to accurately segment certain classes in some specific scenarios, like the ones depicted in this figure. However, these pseudo-labels get filtered during training using the consistency measure with the predictions given by the student model.}
    \label{fig:failure_qualitative_comparison}
\end{figure*}